\documentclass[12pt]{article}

\usepackage[top=1in,bottom=1in,left=1in,right=1in]{geometry}
\usepackage[colorlinks, linkcolor=blue, citecolor=blue]{hyperref}           
\usepackage{graphicx,dsfont}
\usepackage{subcaption}
\usepackage{amsmath}
\usepackage{textcomp, gensymb}
\usepackage[numbers, square, sort&compress]{natbib} 
\usepackage{tikz}
\usepackage{pgfplots}
\usepackage{mathrsfs}  
\usepackage{amssymb}
\usepackage{cancel}
\usepackage{tabularx} 
\usepackage{enumitem}
\usepackage{physics}
\usepackage{import}
\usepackage{pdfpages}
\usepackage{mathtools}
\usepackage{xparse}
\usepackage{bm}
\usepackage{amsthm}
\usepackage[colorlinks, linkcolor=blue, citecolor=blue]{hyperref}  
\usepackage{amsthm,wrapfig,comment}

\definecolor{amber}{rgb}{1.0,0.75,0.0}
\definecolor{americanrose}{rgb}{1.0,0.01,0.24}

\makeatletter
\renewcommand{\paragraph}{%
  \@startsection{paragraph}{4}%
  {\z@}{0ex \@plus 1ex \@minus .2ex}{-1em}%
  {\normalfont\normalsize\bfseries}
}
\makeatother

\usepackage{algorithm}
\usepackage[noend]{algpseudocode}

\newcommand{\cmmnt}[1]{\ignorespaces}

\newcommand{\mor}[1]{\textcolor{blue}{[Mor: #1]}}

\newcommand{\Rmnum}[1]{\expandafter\@slowromancap\romannumeral #1@}
\newcommand{\s}[1]{\mathsf{#1}}
\newcommand{\p}[1]{\left(#1\right)}
\newcommand{\pp}[1]{\left[#1\right]}
\newcommand{\ppp}[1]{\left\{#1\right\}}

\newcommand{\calG}{\mathcal{G}} 
\newcommand{\calV}{\mathcal{V}} 
\newcommand{\calE}{\mathcal{E}} 
\newcommand{\calP}{\mathcal{P}} 
\newcommand{\calH}{\mathcal{H}} 
\newcommand{\calW}{\mathcal{W}} 
\newcommand{\calI}{\mathcal{I}} 
\newcommand{\calF}{\mathcal{F}} 
\newcommand{\calS}{\mathcal{S}} 
\newcommand{\calA}{\mathcal{A}} 
\newcommand{\calB}{\mathcal{B}} 
\newcommand{\calT}{\mathcal{T}} 
\newcommand{\calD}{\mathcal{D}}

\newcommand{\calZ}{\mathcal{Z}}

\newcommand*{\SPRT}[1]{\textnormal{SPRT}\left(#1\right)}

\newtheorem{theorem}{Theorem}

\newtheorem{proposition}{Proposition}

\newenvironment{fminipage}%
  {\begin{Sbox}\begin{minipage}}%
  {\end{minipage}\end{Sbox}\fbox{\TheSbox}}

\NewDocumentCommand{\E}{e{_}e{^}g}{%
	\operatorname{\mathbb{E}}%
	\IfValueT{#1}{_{#1}}
	\IfValueT{#2}{^{#2}}%
	\IfValueT{#3}{\pp{#3}}%
}
\RenewDocumentCommand{\P}{e{_}e{^}g}{%
	\operatorname{\mathbb{P}}%
	\IfValueT{#1}{_{#1}}
	\IfValueT{#2}{^{#2}}%
	\IfValueT{#3}{\p{#3}}%
}

\RenewDocumentCommand{\var}{e{_}e{^}g}{%
	\operatorname{\mathbb{V}\textnormal{ar}}%
	\IfValueT{#1}{_{#1}}
	\IfValueT{#2}{^{#2}}%
	\IfValueT{#3}{\p{#3}}%
}

\makeatletter
\def\thanks#1{\protected@xdef\@thanks{\@thanks
        \protect\footnotetext{#1}}}
\makeatother

\allowdisplaybreaks

\begin{document}
\title{Online Auditing of Information Flow}
\author{Mor~Oren-Loberman~~~~~~~~~Vered Azar~~~~~~~~~Wasim Huleihel\thanks{M. Oren-Loberman, V. Azar and  W. Huleihel are with the Department of Electrical Engineering-Systems at Tel Aviv university, {T}el {A}viv 6997801, Israel (e-mails:  \texttt{orenmor@mail.tau.ac.il, vered.azr@gmail.com, wasimh@tauex.tau.ac.il}). This work is supported by the ISRAEL SCIENCE FOUNDATION (grant No. 1734/21).}}
\maketitle
%

\begin{abstract}
Modern social media platforms play an important role in facilitating rapid dissemination of information through their massive user networks. Fake news, misinformation, and unverifiable facts on social media platforms propagate disharmony and affect society. In this paper, we consider the problem of online auditing of information flow/propagation with the goal of classifying news items as fake or genuine. Specifically, driven by experiential studies on real-world social media platforms, we propose a probabilistic Markovian information spread model over networks modeled by graphs. We then formulate our inference task as a certain sequential detection problem with the goal of minimizing the combination of the error probability and the time it takes to achieve correct decision. For this model, we find the optimal detection algorithm minimizing the aforementioned risk and prove several statistical guarantees. We then test our algorithm over real-world datasets. To that end, we first construct an offline algorithm for learning the probabilistic information spreading model, and then apply our optimal detection algorithm. Experimental study show that our algorithm outperforms state-of-the-art misinformation detection algorithms in terms of accuracy and detection time.
\end{abstract}

\section{Introduction} 
\label{sec:intro}
\allowdisplaybreaks
Modern social media platforms significantly facilitate the rapid dissemination of information through their massive user networks. Recent surveys indicate that 73\% of people receive news from social media \cite{Elisa18}, and as many as 72\% of adult Internet users in the U.S. have used social network sites for health related advice, the majority of which following experiences shared by friends on social media. The ease of posting and sharing news, coupled with recent advances in AI technology, expedite the propagation of rumors, misinformation, and even maliciously fake information. Unfortunately, the ability of AI algorithms to identify such items grows more slowly than the ability to create it \cite{Paschen19}. It is therefore of major importance to develop a better theoretical understanding of the social structure that enables the propagation of such items, to guide the development of robust methodologies and efficient countermeasures to tackle this problem. 

Recently, it was shown empirically in \cite{Vosoughi18} that in many social networks \emph{falsehood information}/\emph{misinformation}/\emph{rumors} diffuse significantly farther, faster, deeper, and more broadly than the \emph{truth}, in all categories of information, and in many cases by an order of magnitude. Moreover, it was observed that the spread of fake information is essentially not due to social bots that are programmed to disseminate inaccurate stories. Instead, fake news speeds faster around, say, Twitter, due to people retweeting inaccurate news items. This observation was later used in \cite{monti2019fake} to provide a heuristic recovery algorithm for fake news detection using geometric deep learning. 

Despite the fact that the topic of automatic misinformation detection received a significant attention in the literature, it is still considered as a challenging daunting task. As an initial approach of combating the spread of misinformation, many social media platforms exploit their massive crowd and resources to employ ``human-based'' methods, such as crowd sourcing user feedbacks and third-party fact-checking. Even though these methods have the potential of achieving high accuracy rate, they are most often unscalable and significantly slow. In order to cope with the crucial drawbacks of these primitive methods, significant work has gone into research on automatic misinformation detection in a fast, scaleable and accurate manner. It has been demonstrated in \cite{castillo2011information} that utilizing users and posts content features for misinformation detection can be very effective. This observation drives many machine learning, data-mining, and AI approaches to automatically detect misinformation using feature extraction (see, a recent survey in \cite{10.1145/3137597.3137600}).

Perhaps surprisingly, in spite of the crucial importance of \emph{quick} misinformation detection due to its deceptive nature, this aspect is still in its early stage of development. Several early misinformation detection algorithms that aim to debunk rumors at their stage of diffusion have been developed \cite{ma2015detect, liu2018early, ruchansky2017csi, chen2018call, 10.5555/3061053.3061153, ma-etal-2017-detect, ma-etal-2018-rumor, zhao2015enquiring}, yet, these do not make a real-time decision and require a pre-determined number of observations as input. 
In this paper, our goal is to provide a \emph{systematic theoretic} investigation of the interesting findings in \cite{Vosoughi18}. In particular, we would like to understand and answer the following meta-question:

\vspace{0.2cm}
\noindent\fbox{\parbox{0.97\textwidth}{
\emph{Is it possible to infer whether pieces of information propagated over time in a social network are falsehood or truth \textbf{based only on the way} they diffuse over the network?}
}}
\vspace{0.2cm}

An intriguing framework was suggested in \cite{wei2019quickstop} for real-time quickest misinformation detection, using a Markov optimal stopping problem, based on a probabilistic information spreading model. Under this framework, the authors propose a data-driven and model-driven algorithm, termed QuickStop, for real-time misinformation detection. This algorithm consists of an offline procedure for learning the probabilistic information spreading model, and an online algorithm to detect misinformation. Our paper is greatly inspired by \cite{wei2019quickstop}, however, we propose and analyze a more general and realistic diffusion model, that takes into account several practical aspects, such as the social network graph structure, and the possibility of missing data, as we explain below in detail. 

Specifically, to approach the question above, we introduce a Markovian information spreading model over a social network modeled by a graph. When a complete network and information diffusion information are known, the information spreading trace is likely to be a tree or a forest (when multiple information sources exist). However, in practice, it is often not the case because of missing information and partial observations, see, e.g., \cite{Jin13,Kwon13}. Accordingly, in our model we assume that only arbitrary parts of the information spreading traces are observed by the auditor. Thus, while the underlying information traces are Markovian, the actual observed information behaves as a more complicated hidden Markov chain. With this model, we define the auditor's goal using a sequential hypothesis testing problem, where under the null hypothesis the underlying information is genuine, and under the alternative hypothesis it is fake. We show that this problem can be formulated as a certain \textit{optimal stopping problem}. 

Using this formulation, we derive an optimal procedure for real-time, quickest cost-efficient misinformation detection, with a straightforward stopping policy. We analyze the performance of this algorithm, by driving bounds on the associated conditional error probabilities, using an equivalent sequential probability ratio test (SQRT) of our detection algorithm. Finally, we test our algorithm over real-world datasets. To that end, we construct an offline algorithm for learning the probabilistic information spreading model, and then apply our optimal detection algorithm. Our experimental results on real-world dataset show that our algorithm outperforms state-of-the-art misinformation detection algorithms in terms of accuracy and detection time.

\paragraph{Notation.}
We use calligraphic font to indicate sets, and sans serif font with uppercase and lowercase letters $\s{X}$, $\s{x}$ to indicate RVs and their values, respectively. $\P(\cdot)$ and $\E{\cdot}$ indicate the probability and expectation functions. $\mathds{1}_{E}$ is the indicator function that gets $1$ when an event $E$ is true and $0$ otherwise. We denote the cardinality of some set $\calS$ by $\abs{\calS}$. For a set $\mathcal{X}$, we let $\mathcal{X}^n$ denote the $n$-fold Cartesian product of $\mathcal{X}$. An element of $\mathcal{X}^n$ is denoted by $x^n =(x_1,x_2,\ldots,x_n)$. A substring of $x^n\in\mathcal{X}^n$ is designated by $x_i^j = (x_i, x_{i+1},\ldots,x_j)$, for $1\leq i \leq j \leq n$; when $i = 1$, the subscript is omitted. A directed walk is a finite or infinite sequence of edges directed in the same direction which joins a sequence of vertices. Let $\calG = (\calV, \calE)$ be a directed graph. A finite directed walk is a sequence of edges $e_1,e_2,\ldots,e_{n-1}$ for which there is an associated sequence of vertices $(v_1,v_2,\ldots,v_n)$ such that $e_i = (v_i,v_{i+1})$, for $i = 1,2,\ldots,n-1$. The sequence $(v_1,v_2,\ldots,v_n)$ is the vertex sequence of the directed walk. A directed path is a directed walk in which all vertices and edges are distinct.

\section{Problem Formulation} \label{section:StatisticalModel}

\paragraph{Underlying graph.} Consider an online social network platform that is monitoring the spread of some information in the network.
Let $\calG = (\calV, \calE)$ be a directed graph representing a social network platform that is monitoring the spread of some information, where $\calV = [n]$ is the set of nodes (e.g., users), and $\calE$ is the set of directed edges (e.g., connections between users). Each node $u\in\calV$ is associated with a $d$-dimensional feature vector $\mathbf{x}_u\in \mathbb{R}^d$. We assume two types of information, either genuine or fake, can spread over $\calG$; we denote its type by $\calI\in\{0,1\}$, with $\calI=0$ refers to a genuine information, and $\calI=1$ refers to a fake one. Given a directed edge $e = (u,v)$, we shall refer to user $v$ as a \emph{follower} of user $u$, while user $u$ is a \emph{followee} of user $v$. User $v\in\calV$ can forward information $\calI$ from user $u \in\calV$. User $u$ decides whether to spread some information or not based on: (i) the information type $\mathcal{I} \in \{0, 1\}$; (ii) the features $\mathbf{x}_u$ of user $u$; and (iii) the set of its neighbors $\mathcal{N}_u \triangleq \{v \in \calV: (u , v) \in \calE\}$, who forwarded the information earlier.

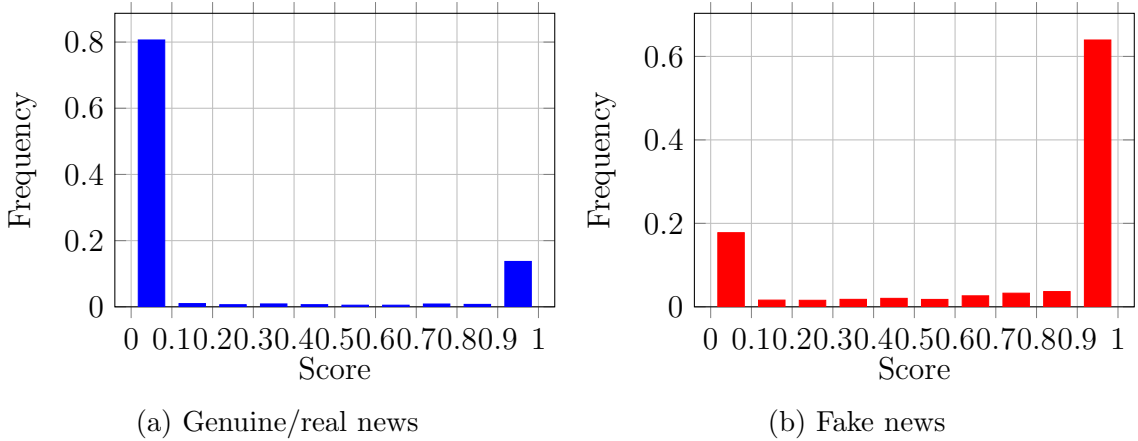
\begin{figure*}[!t]
    \centering

    \begin{subfigure}{.45\linewidth} 
        \centering
        \begin{tikzpicture}
            \begin{axis}[
                ybar, 
                bar width=10,
                xlabel={Score},
                ylabel={Frequency},
                ymin=0,
                width=1.0\textwidth,
                height=.24\textheight,
                xtick={0, 0.1, 0.2, 0.3, 0.4, 0.5, 0.6, 0.7, 0.8, 0.9, 1},
                xticklabels={0, 0.1, 0.2, 0.3, 0.4, 0.5, 0.6, 0.7, 0.8, 0.9, 1},
                grid,
            ]
                \addplot[color=blue, fill=blue] coordinates {(0.05000009500000795, 0.8063008091184196) (0.15000008500000395, 0.009705102858485979) (0.25000007499999993, 0.006363422685248568) (0.3500000649999959, 0.008668740373041425) (0.45000005499999196, 0.006426723234785664) (0.550000044999988, 0.004876105122094002) (0.6500000349999839, 0.0048711981027500404) (0.7500000249999799, 0.008426333617449754) (0.850000014999976, 0.007298700572207526) (0.950000004999972, 0.13706286431551742)};
            \end{axis}
        \end{tikzpicture}
        \caption{Genuine/real news}
        \label{fig:classification_dist_news}
    \end{subfigure}~
    \begin{subfigure}{.45\linewidth} 
        \centering
        \begin{tikzpicture}
            \begin{axis}[
                ybar, 
                bar width=10,
                xlabel={Score},
                ylabel={Frequency},
                ymin=0,
                width=1\textwidth,
                height=.24\textheight,
                xtick={0, 0.1, 0.2, 0.3, 0.4, 0.5, 0.6, 0.7, 0.8, 0.9, 1},
                xticklabels={0, 0.1, 0.2, 0.3, 0.4, 0.5, 0.6, 0.7, 0.8, 0.9, 1},
                grid,
            ]
                \addplot[color=red, fill=red] coordinates {(0.05000009500000795, 0.17762230381254665)
                (0.15000008500000395, 0.016056002371260137)(0.25000007499999993, 0.015614367074719693) (0.3500000649999959, 0.01788563431407055) (0.45000005499999196, 0.020262449611714087) (0.550000044999988, 0.01770826596047344) (0.6500000349999839, 0.026334240544024848) (0.7500000249999799, 0.032660378488988284) (0.850000014999976, 0.036497010728206214) (0.950000004999972, 0.6393593470939961)};
            \end{axis}
        \end{tikzpicture}
        \caption{Fake news}
        \label{fig:classification_dist_fake}
    \end{subfigure}
    
    \caption{Distributions of linear SVM classification scores, associated with the edge-based model, over the \texttt{Weibo} dataset.}
    \label{fig:SVMclassification}
\end{figure*}

\paragraph{Edge types.} 
In this paper, we say that an event occurs when a user (follower) forwards the information from one of its followees. This is represented by the edge over which this event occurs, which in turn is represented by a pair of features $(\mathbf{x}_u,\mathbf{x}_v)$ that correspond to the end users; 
Our edge-based model views each edge $e\in \calE$ as a communication channel associated with a given weight. Specifically, we classify the edges into $\s{Z}\in\mathbb{N}$ types, so that each edge $e\in\calE$ has an associated weight $W_{u,v}\triangleq W_e\in\calZ$, where $\calZ\triangleq\{0,1,\ldots,\s{Z}-1\}$. An edge with a larger weight is more likely to spread misinformation, i.e., an edge with weight $0$ is the type of edges that are more likely to spread genuine information, while an edge with weight $\s{Z}-1$ is more likely to spread fake information. 
Accordingly, we assume that $W_{u,v} = f(\mathbf{x}_u , \mathbf{x}_v)$, where $f:\calE\to\calZ$ is an edge-classifier function. Finally, we denote by $\mathbf{W}$ the $\abs{\calE} \times \abs{\calE}$ matrix with $[\mathbf{W}]_{u,v} = W_{u,v}$, for any $(u,v)\in\calE$. As will be explained later on, our decision algorithm will be based on partial observations of these weights, i.e., retweets are represented by the weights. 
Fig.~\ref{fig:SVMclassification} demonstrates the strong correlation between the information type $\calI$ and the SVM classification score of the edge-based model on the \texttt{Weibo} dataset \cite{10.5555/3061053.3061153}. More specifically, Fig.~\ref{fig:classification_dist_news} illustrates that the scores of edges involved in spreading genuine information are concentrated around zero, while Fig.~\ref{fig:classification_dist_fake} shows that the scores of edges involved in spreading fake information are concentrated around one. This is consistent with the fact that an edge with a higher weight is more likely to spread fake information.

\begin{figure*}[t!]
    \centering
    \begin{tikzpicture}[
        every node/.style={circle, draw, thin, minimum size=8mm, scale=0.7},
        grow=down,
        level distance=1.4cm,
        edge from parent/.style={->, -latex, draw},
        edge/.style = {->, -latex, draw},
        level 1/.style={sibling distance=50mm},
        level 2/.style={sibling distance=17mm},
        level 3/.style={sibling distance=6.5mm}
    ]
    \node (1) {1}
        child {node (2) {2}
            child {node (5) {5}
                child {node (12) {12}}
                child {node (13) {13}}
                child {node (14) {14}}
            }
            child {node (6) {6}
                child {node (15) {15}}
            }
        }
        child {node (3) {3}
            child {node (7) {7}
                child {node (16) {16}}
                child {node (17) {17}}
                child {node (18) {18}}
            }
            child {node (8) {8}
                child {node (19) {19}}
                child {node (20) {20}}
            }
            child {node (9) {9}
                child {node (21) {21}}
                child {node (22) {22}}
            }
        }
        child {node (4) {4}
            child {node (10) {10}
                child {node (23) {23}}
                child {node (24) {24}}
            }
            child {node (11) {11}
                child {node (25) {25}}
                child {node (26) {26}}
                child {node (27) {27}}
            }
        }
        ;  
        \draw[edge] (1) to (6);
        \draw[edge] (6) to (14);
        \draw[edge] (6) to (16);
        \draw[edge] (7) to (19);
        \draw[edge] (9) to (23);
        \draw[edge] (4) to (24);
        \draw[edge] (11) to (27);
        \draw[edge] (1) to (23););

        \draw[edge, white] (1) -- (2);
        \draw[edge, white] (2) -- (6);
        \draw[edge, white] (6) -- (14);
        \draw[edge, white] (1) -- (3);
        \draw[edge, white] (3) -- (7);
        \draw[edge, white] (7) -- (19);
        \draw[edge, white] (1) -- (23);
        \draw[edge, white] (1) -- (4);
        \draw[edge, white] (4) -- (24);

        \draw[edge, red, very thick, dashed] (1) -- (2)  node[midway, left, yshift=4pt, draw=none] {$\s{W}^{\s{P}_1}_{2,1}$};
        \draw[edge, red, very thick, dashed] (2) -- (6) node[midway, right, yshift=4pt, xshift=-6pt, draw=none] {$\s{W}^{\s{P}_1}_{6,2}$};
        \draw[edge, red, very thick, dashed] (6) -- (14) node[midway, left, yshift=8pt, xshift=9pt, draw=none] {$\s{W}^{\s{P}_1}_{14,6}$};
        
        \draw[edge, very thick, blue] (1) -- (3) node[midway, left, xshift=6pt, draw=none] {$\s{W}^{\s{P}_2}_{3,1}$};
        \draw[edge, very thick, blue] (3) -- (7) node[midway, left, yshift=4pt, xshift=8pt, draw=none] {$\s{W}^{\s{P}_2}_{7,3}$};
        \draw[edge, very thick, blue] (7) -- (19) node[midway, right, yshift=7pt, xshift=-8pt, draw=none] {$\s{W}^{\s{P}_2}_{19,7}$};

        \draw[edge, orange, very thick, dotted] (1) -- (23) node[midway, right, draw=none] {$\s{W}^{\s{P}_3}_{23,1}$};
        
        \draw[edge, green, very thick, dashdotted] (1) -- (4) node[midway, right, yshift=4pt, draw=none] {$\s{W}^{\s{P}_4}_{4,1}$} ;
        \draw[edge, green, very thick, dashdotted] (4) -- (24) node[midway, right, yshift=4pt, xshift=-4pt, draw=none] {$\s{W}^{\s{P}_4}_{24,4}$};
    \end{tikzpicture}
    \caption{A partial social media graph with a single information source at $s = 1$. Each weighted path in the graph corresponds to a different Markov chain.}
    \label{Fig:GraphExmp}
\end{figure*}
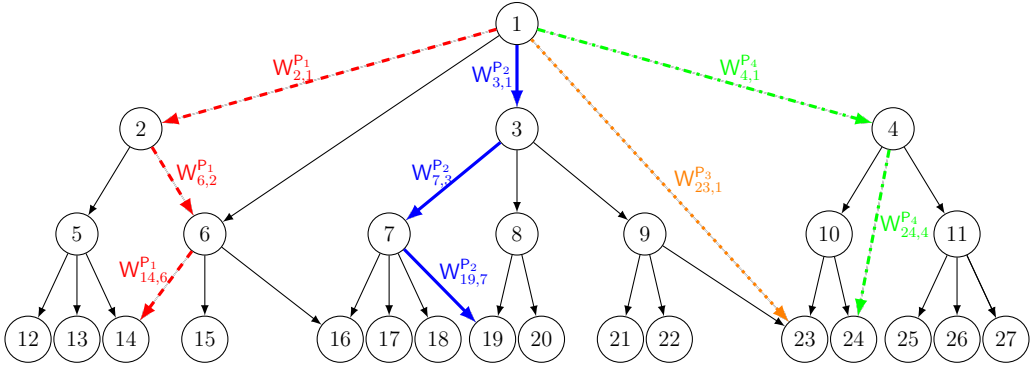

\paragraph{Probabilistic information flow.} Given the above setup for the underlying social network, and edge types, we next define the way information flow/spread over the network. 
We focus on a single information source $s\in\calV$, and assume the following probabilistic model. 
Let $\calP$ denote the set of all possible directed paths in $\calG$ starting at $s$. The information $\calI$ can flow over one or more of these paths. 
As mentioned above, an event occurs when a user retweets the information. Accordingly, for each possible path in $\calG$ we model the sequence of \emph{consecutive} retweets (or, equivalently edges) as a \emph{first-order Markov chain}. Mathematically, for any path $\s{P}\in\calP$ in $\calG$, let its vertex sequence be denoted by $\s{P} = (v_{1}^{\s{P}},v_{2}^{\s{P}},\ldots,v_{|\s{P}|}^{\s{P}})$. Then, each such path is associated with a first-order homogeneous Markov chain $\s{W}^{\s{P}} \triangleq \{\s{W}_{v_{i}^{\s{P}},v_{i+1}^{\s{P}}}\}_{i=1}^{|\s{P}|-1}$ with $\s{Z}$ states. In particular, each retweet of $\calI$ by a followee-follower pair $(u,v)\in\s{P}$ is represented by the weight (edge-type) $\s{W}^\s{P}_{u,v}$, which in turn corresponds to a feature pair $(\mathbf{x}_u,\mathbf{x}_v)$, that is classified into one of $\s{Z}$ states by $f(\cdot)$, as described above. 
Given $\calI$, we let $\alpha_{\calI}(\s{z}\vert\s{z}')$, for $\s{z},\s{z}'\in\calZ$, denote the edge transition probabilities that depend on the underlying information type. 
These transition probabilities clearly depend on whether the underlying information is genuine or not. Indeed, it is reasonable that $\alpha_{0}(0 \vert 0) > \alpha_{1}(0 \vert 0)$, namely, the transition between two edges of type ``$0$" is more probable if a genuine message is being propagated. The motivation is that if the transition probabilities $\alpha_0(\cdot \vert \cdot)$ and $\alpha_1(\cdot \vert \cdot)$ are ``sufficiently distinctive", we should be able decide whether genuine or fake information is propagated. Furthermore, for any $u\in\mathcal{N}_s$ we define the initial probabilities $\eta_{\calI}(\s{z})\triangleq\P(\s{W}_{s,u}=\s{z}\vert\calI)$, for any $\s{z}\in\calZ$, and $\calI\in\{0,1\}$. Later on, we will devise a data-based offline algorithm to learn all the above terms. Tables \ref{tab:trans_prob} and \ref{tab:init_prob} show the empirical transition probabilities $\alpha_0(\cdot \vert \cdot)$ and $\alpha_1(\cdot \vert \cdot)$, and the initial probabilities $\eta(\cdot)$, estimated using the \texttt{Weibo} dataset, with $\s{Z}=4$, respectively. Throughout this paper, we follow the experimental setting in Section~\ref{section:Experiments}. 
Thenceforth, with some abuse of notation, for a given path $\s{P}\in\calP$, we let $\s{W}_{v\to u}^{\s{P}}$ denote the trajectory of edge weights starting from user $v\in\calV$ and ending at user $u\in\calV$. 
Fig.~\ref{Fig:GraphExmp} gives an example of a social media graph with multiple Markov chains paths. 

\begin{table*}[ht!]
    \begin{subtable}{0.45\linewidth}
    \centering
    \begin{tabular}{|c|*{4}{>{\centering\arraybackslash}p{1cm}|}}
        \hline
        \  & 0 & 1 & 2 & 3 \\
        \hline
        0 & 0.159 & 0.029 & 0.191 & 0.621 \\
        \hline
        1 & 0.959 & 0.001 & 0.001 & 0.039 \\
        \hline
        2 & 0.057 & 0.017 & 0.016 & 0.91 \\
        \hline
        3 & 0.057 & 0.145 & 0.027 & 0.771 \\
        \hline
    \end{tabular}
    \caption{Real news}
    \end{subtable}
    \hfill
    \begin{subtable}{0.45\linewidth}
    
    \centering
    \begin{tabular}{|c|*{4}{>{\centering\arraybackslash}p{1cm}|}}
        \hline
        \  & 0 & 1 & 2 & 3 \\
        \hline
        0 & 0.659 & 0.017 & 0.028 & 0.297 \\
        \hline
        1 & 0.065 & 0.015 & 0.021 & 0.899 \\
        \hline
        2 & 0.064 & 0.011 & 0.075 & 0.85 \\
        \hline
        3 & 0.026 & 0.004 & 0.006 & 0.964 \\
        \hline
    \end{tabular}
    \caption{False news}
    \end{subtable}
    \caption{Edge transition probability matrices $\alpha(\cdot \vert \cdot)$ from the \texttt{Weibo} dataset, for $\s{Z}=4$.}
    \label{tab:trans_prob}
\end{table*}

\begin{table*}[ht!]
    \centering
    \begin{tabular}{|c|c|c|c|c|}
    \hline
    \  & 0 & 1 & 2 & 3 \\
    \hline
    Real News & 0.872 & 0.004 & 0.003 & 0.120 \\
    \hline
    False News & 0.101 & 0.006 & 0.015 & 0.876 \\
    \hline
    \end{tabular}
    \caption{Edge initial probabilities $\eta(\cdot)$ from the \texttt{Weibo} dataset, for $\s{Z}=4$.}
    \label{tab:init_prob}
\end{table*}

\begin{figure*}[t!]
    \centering
    \begin{tikzpicture}[
        every node/.style={circle, draw, thin, minimum size=8mm, scale=0.7},
        grow=down,
        level distance=1.4cm,
        edge from parent/.style={->, -latex, draw},
        edge/.style = {->, -latex, draw},
        level 1/.style={sibling distance=50mm},
        level 2/.style={sibling distance=17mm},
        level 3/.style={sibling distance=6.5mm}
    ]
    \node (1) {1}
        child {node (2) {2}
            child {node (5) {5}
                child {node (12) {12}}
                child {node (13) {13}}
                child {node (14) {14}}
            }
            child {node (6) {6}
                child {node (15) {15}}
            }
        }
        child {node (3) {3}
            child {node (7) {7}
                child {node (16) {16}}
                child {node (17) {17}}
                child {node (18) {18}}
            }
            child {node (8) {8}
                child {node (19) {19}}
                child {node (20) {20}}
            }
            child {node (9) {9}
                child {node (21) {21}}
                child {node (22) {22}}
            }
        }
        child {node (4) {4}
            child {node (10) {10}
                child {node (23) {23}}
                child {node (24) {24}}
            }
            child {node (11) {11}
                child {node (25) {25}}
                child {node (26) {26}}
                child {node (27) {27}}
            }
        }
        ;  
        \draw[edge] (1) to (6);
        \draw[edge] (6) to (14);
        \draw[edge] (6) to (16);
        \draw[edge] (7) to (19);
        \draw[edge] (9) to (23);
        \draw[edge] (4) to (24);
        \draw[edge] (11) to (27);
        \draw[edge] (1) to (23););

        \draw[edge, white] (1) -- (2);
        \draw[edge, white] (1) -- (23);
        \draw[edge, white] (3) -- (7);
        \draw[edge, white] (7) -- (19);
        \draw[edge, white] (4) -- (24);
        \draw[edge, white] (6) -- (14);

        \draw[edge, ultra thick, red] (1) -- (2)  node[midway, left, yshift=4pt, xshift=-4pt, draw=none] {$\s{Z}_{1}$};
        \draw[edge, ultra thick, red] (1) -- (23)  node[midway, right, yshift=4pt, xshift=-4pt, draw=none] {$\s{Z}_{2}$};
        \draw[edge, ultra thick, red] (3) -- (7)  node[midway, left, yshift=6pt, xshift=4pt, draw=none] {$\s{Z}_{3}$};
        \draw[edge, ultra thick, red] (7) -- (19)  node[midway, right, yshift=6pt, xshift=-4pt, draw=none] {$\s{Z}_{4}$};
        \draw[edge, ultra thick, red] (4) -- (24)  node[midway, right, yshift=4pt, xshift=-3pt, draw=none] {$\s{Z}_{5}$};
        \draw[edge, ultra thick, red] (6) -- (14)  node[midway, left, yshift=6pt, xshift=4pt, draw=none] {$\s{Z}_{6}$};
        
    \end{tikzpicture}
    \caption{An illustration of a possible sequence of observations $\{\s{Z}_1,\ldots,\s{Z}_{6}\}$ in a social media graph with a single source $s = 1$. 
    }
    \label{Fig:HiddenMarkov}
\end{figure*}
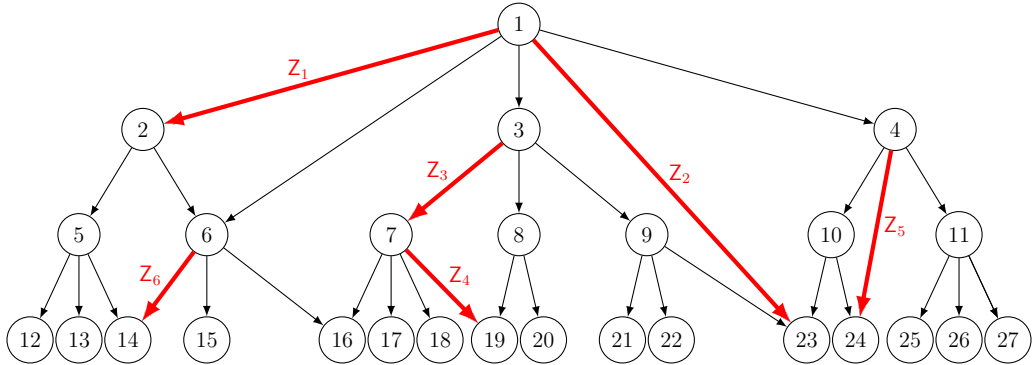

\paragraph{Observations and learning problem.} 
We now formulate the misinformation detection problem as a sequential hypothesis testing problem. Firstly, we define the type of observations available for testing. 
When complete network and diffusion information are known, the information spreading trace forms a tree with its paths forming Markov chains. Alas, in practice, it is often not the case due to missing information and partial observations \cite{Jin13,Kwon13}, thus, we assume that we observe only arbitrary parts of the information spreading traces which we denote by the sequence $\{\s{Z}_{\ell}\}_{\ell\geq1}$, where $\s{Z}_{\ell} = \s{W}_{u,v}^{\s{P}}$, for some path $\s{P}\in\calP$, and edge $(u,v)\in\calE$.
Given $\calI$, the sequence $\{\s{Z}_{\ell}\}_{\ell\geq1}$ is subjected to a joint probability law that is governed by the transition probabilities $\alpha_{\calI}(\cdot\vert\cdot)$. Below, when the underlying information is genuine (fake) $\calI=0$ ($\calI=1$), we say that $\{\s{Z}_{\ell}\}_{\ell\geq1}$ is generated from $\mathcal{L}_0$ ($\mathcal{L}_1$). An illustration of this observation model is given in Fig.~\ref{Fig:HiddenMarkov}. Compared to Fig.~\ref{Fig:GraphExmp}, it can be seen that the actual observations are subsamples of some edges on the complete paths.

Our learning problem is formulated as follows. Consider a sequence $\{\s{Z}_{\ell}\}_{\ell\ge1}$ that obey one of the two hypotheses, and the audit is tasked with distinguishing between the two hypotheses,
\begin{align}\label{eqn:learning}
    \calH_0:\calI=0\quad\quad\s{vs.}\quad\quad\calH_1:\calI=1,
\end{align}
namely, the underlying information is genuine (null hypothesis) or fake (alternative hypothesis). In terms of the prior distribution, we assume that hypothesis $\calH_1$ occurs with probability $\pi_1 = \pi$, and $\calH_0$ occurs with probability $\pi_0 \triangleq 1 - \pi$, for some $\pi\in[0,1]$. The audit is tasked with distinguishing between the two hypotheses in a way that minimizes a combination of the error probability and the propagation cost, as we define in the sequel. Consider a probability space $(\Omega,\calF,\mathbb{P}_\pi)$, where $\P_{\pi}$ is the probability measure defined as follows
\begin{align}
    \P_{\pi} = (1 - \pi) \P_0 + \pi \P_1,
\end{align}
with $\P_0$ and $\P_1$ being the probability measures under the null and alternative hypotheses, respectively, such that, under $\P_{\calI}$, the sequence $\{\s{Z}_{\ell}\}_{\ell\ge1}$ is generated from $\mathcal{L}_{\calI}$, for $\calI=\{0,1\}$. Now, note that the measures $\P_0$ and $\P_1$ are mutually singular, since the likelihood ratio yields
\begin{align}
    \Lambda_{\ell} \triangleq \frac{\P_1(\s{Z}_1,\ldots,\s{Z}_\ell)}{\P_0(\s{Z}_1,\ldots,\s{Z}_\ell)}
        \xrightarrow{\ell\to\infty} 
        \begin{cases}
            0, & \textnormal{a.s. under } \P_0\\
            \infty, & \textnormal{a.s. under } \P_1.
        \end{cases} \label{eq:distingushing}
\end{align}
That is, we can tell the distributions apart from the limiting value of the likelihood ratio. So, if we observe $\{\s{Z}_{\ell}\}_{\ell\ge1}$ we can decide perfectly between the two hypotheses. In this paper, however, there is an additional cost for sampling. Specifically, suppose we observe $\{\s{Z}_{\ell}\}_{\ell\ge1}$ sequentially, generating the natural filtration $\{\calF_{\ell}\}_{\ell\ge1}$, with 
\begin{align}
    \calF_{\ell} \triangleq \{\s{Z}_1, \s{Z}_2, \ldots , \s{Z}_{\ell}\},
\end{align}
and $\calF_0 \triangleq (\Omega, \emptyset)$. Clearly, a tradeoff develops between the decision accuracy and the potential damage of spreading misinformation. The arising optimization problem can be examined in the context of \textit{sequential decision rule}. 
Let $\s{T} \in \calT$ be the random stopping time at which the type of information is declared, where $\calT$ denotes the set of all stopping times $\s{T} \ge 1$ with respect to the filtration $\ppp{\calF_{\ell}}$, and a sequence $\ppp{\delta_{\ell}}$ of terminal decision rules, where $\delta_{\ell}$ is an $\calF_{\ell}$-measure function taking the values $\ppp{0,1}$. Let $\calD$ denote the set of all such $\delta$. Then, a sequential decision rule is defined as 
\begin{align}
    \delta_{\s{T}} \triangleq \sum_{\ell = 0}^{\infty} \delta_{\ell} \mathds{1}_{\{ \s{T} = \ell \}},    
\end{align}
where $\mathds{1}_{\{ \s{T} = \ell \}}$ is the indicator function that gets $1$ when $\ell$ is the stopping time $\s{T}$ and $0$ otherwise. Obviously, $\delta_{\s{T}}$ is an $\calF_{\s{T}}$-measure as well and also taking the values $\ppp{0, 1}$. 
Now, a sequential decision rule is described as $(\s{T},\delta)$, in which $\s{T}$ declares the time to stop sampling, and once $\s{T}$ is given, $\delta_{\s{T}}$ takes the values $0$ or $1$ declaring which hypotheses to accept: $\calH_0$ for genuine information and $\calH_1$ for fake information, correspondingly. 
For a given $(\s{T},\delta)$, we define the average cost of errors due to misdetection, as
    \begin{align}
        c_{e}(\s{T}, \delta_{\s{T}}) 
            \triangleq  c_{\s{I}}
                \P_{\calH_0}(\delta_{\s{T}} = 1) 
                + c_{\s{II}}
                \P_{\calH_1}(\delta_{\s{T}} = 0),
    \end{align}
    where $c_{\s{I}}$, and $c_{\s{II}}$ are the costs of type-I error (news is declared as misinformation) and type-II error (misinformation is declared as news). Furthermore, the \textit{propagation cost} due to spreading misinformation is given by $c \E[\s{T} \mathds{1}_{\calH_1}]$, where $c \in \mathbb{R}_+$ is the cost of spreading misinformation at each time slot, and $\mathds{1}_{\calH_1}$ is the indicator function that gets the value 1 when hypothesis $\calH_1$ is true and 0 otherwise. Our main goal is to find the stopping time $\s{T}$ and decision rule $\delta$ that minimize the combination of the error probability and propagation costs, i.e.,
\begin{align} 
    \inf\limits_{\mathsf{T} \in \calT,\delta\in\calD} c_e(\s{T}, \delta_\s{T}) + c\E\pp{\s{T}  \mathds{1}_{{\calH_1}}}.\label{eq:OptimalStoppingProblem}
\end{align}
In the following section, we derive closed-form expressions for the optimal detector and stopping rule, along with several performance guarantees. We then propose model and data driven algorithms, which in addition to the above, classify the edge types and estimate the transition probabilities.

\section{Main Results} \label{section:AlgoAndProformance}

\subsection{Optimal sequential test}
In this subsection, we find the optimal test minimizing the objective function in \eqref{eq:OptimalStoppingProblem}. To that end, we start by presenting a few definitions. Let 
\begin{align}
    \Pi_{\ell} 
        &\triangleq \P{\calH_1 \vert \calF_{\ell}},
\end{align}
denote the posterior probability. Also, let $A_{\calI} (\s{Z}_{\ell} \vert \calF_{\ell -1})$ denote the conditional probability of an observation $\s{Z}_{\ell}$ given all the previous observations and the hypothesis $\calH_{\calI}$, namely, 
\begin{align}
    A_{\calI}(\s{Z}_{\ell} \vert \calF_{\ell-1}) 
        &\triangleq \P{\s{Z}_{\ell} \vert \calF_{\ell-1}, \calH_{\calI}}.
\end{align}
Let $\calP_{\ell}$ denote the set of all directed paths leading to $\s{Z}_{\ell}$ from the source $s$. For each path $\s{P} \in \calP_{\ell}$, we let $\s{Z}^{\s{P}}_{\s{I}_{\ell}(\s{P})}$ denote the last observation in the sequence $(\s{Z}_1, \ldots, \s{Z}_{\ell - 1})$ before $\s{Z}_\ell$ in the path. Then, note that the sequence of observations $(\s{Z}^{\s{P}}_{\s{I}_{\ell}(\s{P})},\s{W}^{\s{P}}_{\s{I}_{\ell}(\s{P})+1\to\ell-1},\s{Z}^{\s{P}}_{\ell})$ across the path $\s{P}$ form a first-order Markov chain. This is illustrated in Fig.~\ref{Fig:hidden_Markov_exmp}.
\begin{figure} [t!]
    \centering
    \begin{tikzpicture}[
        every node/.style={circle, draw, thin, minimum size=12mm, scale=0.6},
        grow=down,
        level distance=1.4cm,
        edge from parent/.style={->, -latex, draw},
        edge/.style = {->, -latex, draw},
    ]
    \node (1) at (0, 0) {$v_{\s{I}_{\ell}(\s{P})}^{\s{P}}$};
    \node (2) at (1, -1) {$u_{\s{I}_{\ell}(\s{P})}^{\s{P}}$};
    \node (3) at (2, -2) {};
    \node (4) at (1, -3) {};
    \node (5) at (1, -4) {$v_{\ell}$};
    \node (6) at (0, -5) {$u_{\ell}$};

    \draw[edge, red, ultra thick] (1) -- (2)  node[midway, right, yshift=8pt, xshift=-2pt, draw=none] {$\s{Z}^{\s{P}}_{\s{I}_{\ell}(\s{P})} = \s{W}^{\s{P}}_{u_{\s{I}_{\ell}(\s{P})}^{\s{P}}, v_{\s{I}_{\ell}(\s{P})}^{\s{P}}}$};
    \draw[edge] (2) -- (3);
    \draw[edge] (3) -- (4);
    \draw[edge] (4) -- (5);
    \draw[edge, red, ultra thick] (5) -- (6)  node[midway, right, yshift=-4pt, xshift=-2pt, draw=none] {$\s{Z}_{\ell} = \s{W}^{\s{P}}_{u_{\ell},v_{\ell}}$};
    \end{tikzpicture}
    \caption{An illustration of a path $\s{P} \in \calP_{\ell}$ which forms the Markov-chain between $\s{Z}^{\s{P}}_{\s{I}_{\ell}(\s{P})} = \s{W}^{\s{P}}_{u_{\s{I}_{\ell}(\s{P})}^{\s{P}}, v_{\s{I}_{\ell}(\s{P})}^{\s{P}}}$ and $\s{Z}_{\ell} = \s{W}^{\s{P}}_{u_{\ell},v_{\ell}}$.}
    \label{Fig:hidden_Markov_exmp}
\end{figure}
For simplicity of notation, we define the path score probability measure as 
\begin{align}
    \mu_{\calI}(\s{P}) 
        \triangleq \P\p{\s{P} \vert \calF_{\ell - 1}, \calH_{\calI}},\label{eqn:muDef}
\end{align}
for any $\s{P}\in\calP_\ell$, and $\calI\in\{0,1\}$. We show in the proof of Theorem~\ref{thm:recursive} below that $\mu_{\calI}(\s{P})$ is given by \eqref{eqn:defMuIP}, shown at the top of the next page,
\begin{figure*}
\begin{align}
    \mu_{\calI}(\s{P}) = \frac{\prod\limits_{i\in \mathcal{J}_{\ell-1}^{\s{P}}} \sum\limits_{\s{z}_{\s{I}_{i}(\s{P})+1}^{i-1}\in\calZ^{i-\s{I}_{i}(\s{P})-1}}
    \prod\limits_{j=\s{I}_{i}(\s{P})+1}^{i} \alpha_{\calI} (\s{z}_{j}\vert\s{z}_{j-1})}{\sum\limits_{\s{P}'\in\calP_{\ell}}\prod\limits_{i\in\mathcal{J}_{\ell-1}^{\s{P}}}\sum\limits_{\s{z}_{\s{I}_{i}(\s{P}')+1}^{i-1}\in\calZ^{i-\s{I}_{i}(\s{P}')-1}}
    \prod\limits_{j=\s{I}_{i}(\s{P}')+1}^{i} \alpha_{\calI} (\s{z}_{j}\vert\s{z}_{j-1})}\label{eqn:defMuIP}
\end{align}
\end{figure*}
where $\mathcal{J}_{\ell-1}^{\s{P}}$ is a set of indices of the observations in the sequence $(\s{Z}_1, \!\ldots\!, \s{Z}_{\ell-1})$ that belong to the path $\s{P}$. We are now in a position to state our first main result.

\begin{theorem}[Posterior probability recursion]\label{thm:recursive}
    The following recursive relation holds,
    \begin{align}
    &\Pi_{\ell + 1} = 
        \frac{\Pi_{\ell} A_1(\s{Z}_{\ell + 1} \vert \calF_{\ell})} {\Pi_{\ell} A_1(\s{Z}_{\ell+1} \vert \calF_{\ell}) 
        + (1 - \Pi_{\ell}) A_0(\s{Z}_{\ell + 1} \vert \calF_{\ell})}, \label{eq:SimpleFormRecursiveEquation}
\end{align}
where $A_{\calI}(\s{Z}_{\ell} \vert \calF_{\ell-1}) = \eta_\calI(\s{Z}_{\ell})$ if $\s{Z}_\ell$ is connected directly to the source, otherwise, we have \eqref{eq:NextObservationConditionalProbability}, shown at the top of the next page.
\begin{figure*}
\begin{align}\label{eq:NextObservationConditionalProbability}
    A_{\calI}(\s{Z}_{\ell} \vert \calF_{\ell-1})= \E_{\mu_{\calI}}\pp{\sum_{\s{z}_{\s{I}_{i}(\s{P})+1}^{i-1}\in\calZ^{i-\s{I}_{i}(\s{P})-1}}
    \prod_{j=\s{I}_{i}(\s{P})+1}^{i} \alpha_{\calI} (\s{z}_{j}\vert\s{z}_{j-1})}
\end{align}
\end{figure*}
\end{theorem}

Next, we show that the optimization problem in \eqref{eq:OptimalStoppingProblem} can cast as the following optimal stopping problem.
\begin{theorem}[Optimal sequential test] \label{theorem:OptimalStoppingProblem}
The optimal stopping problem in \eqref{eq:OptimalStoppingProblem} is equivalent to the following optimal stopping problem,
    \begin{align}
        \inf_{\mathsf{T},\delta} c_e(\s{T}, \delta_{\s{T}}) + c \E[\s{T}  \mathds{1}_{{\calH_1}}] =\inf\limits_{\s{T}} 
            \E[
                g(\Pi_{\s{T}})
                + c \s{T} \Pi_{\s{T}}], \label{eq:equivallenProb}
    \end{align}
    with $g(\pi) \triangleq \min\{c_{\s{II}}\pi, c_{\s{I}}(1-\pi)\}$.
    Furthermore, the optimal test minimizing \eqref{eq:OptimalStoppingProblem} is given by  $\delta_{\s{T}} = \mathds{1}_{\{c_{\s{II}} \Pi_{\s{T}} > c_{\s{I}} (1 - \Pi_{\s{T}})\}}$.
\end{theorem}
Theorem \ref{theorem:OptimalStoppingProblem} implies that the 
only variable in the equivalent optimal stopping problem is the stopping time $\s{T}$. Furthermore, it can be seen that we converted the problem of minimizing over both the stopping time and the decision rules to the new problem in \eqref{eq:equivallenProb}, where we only need to find the optimal stopping time. Accordingly, we can find the optimal stopping policy in two steps: first find the optimal stopping time $\s{T}$ by solving \eqref{eq:equivallenProb}, and then find the optimal decision rule $\delta_{\s{T}}$. We next characterize the optimal stopping time rule. To that end, we need a few definitions. Let $\calT_{\ell}$ denote the subset of the set of all stopping times $\calT$ with respect to the filtration $\calF_{\ell}$ satisfying $\P(\s{T} \ge \ell) = 1$, for all $\ell\geq 1$. For $\ell = 1,2,\ldots$ we define the sequence $\{s_{\ell}\}_{\ell\geq1}$,
\begin{align}
    s_{\ell}(\pi,\s{z}_1^{\ell})\triangleq \inf \limits_{\s{T} \in \calT_{\ell}} 
            \E[ g(\Pi_{\s{T}}) + 
                c \s{T} \Pi_{\s{T}} 
                \vert 
                \Pi_\ell = \pi, \calF_{\ell} = \s{z}_1^{\ell}
            ],
\end{align}
where $s_0(\pi,\emptyset) \triangleq g(\pi)$. 
Note that $s_{\ell}(\pi, \s{z}_1^{\ell})$ is the minimum expected total cost if the algorithm is obligated to stop at time $\s{T} \ge \ell$, conditioned on the information up to $\ell$. Also, define $\bar{s}_{\ell}\p{\pi, \s{z}_1^{\ell}} 
    =   s_{\ell}\p{\pi, \s{z}_1^{\ell}} 
        - c \ell \pi$, for $\ell\geq1$. 
for $\ell\geq1$. We have the following result.

\begin{theorem}[Optimal stopping time]
    \label{theorem:OptimalStoppingTime}
    The optimal stopping time $\s{T}^\star$, achieving the minimum in \eqref{eq:equivallenProb}, is given by,   
    \begin{align}
        \s{T}^\star = \inf 
                \{
                    \ell\in\mathbb{N} 
                    \!:
                    \Pi_{\ell} \!\notin\! (\pi_{\s{low}}(\s{z}_1^{\ell}) , 
                    \pi_{\s{up}}(\s{z}_1^{\ell}))\},
                \label{eq:optimalStoppingTime}
    \end{align}and the optimal decision rule is
    \begin{align}
        \delta_{\s{T}^\star}(\s{z}_1^{\s{T}^\star}) = 
            \begin{cases}
                0,  &   \Pi_{\s{T}^\star} \le \pi_{\s{low}}(\s{z}_1^{\s{T}^\star}) \\
                1,  &   \Pi_{\s{T}^\star} > \pi_{\s{up}}(\s{z}_1^{\s{T}^\star}),
            \end{cases}
            \label{eq:decision_rule}
    \end{align}
with
    \begin{align}
     \pi_{\s{low}}(\s{z}_1^{\ell}) 
        &= \sup 
            \ppp{
                0 \le \pi \le  \frac{c_{\s{I}}}{c_{\s{I}} + c_{\s{II}}} 
                : \bar{s}_{\ell} \p{\pi, \s{z}_1^{\ell}} 
                    = c_{\s{II}} \pi
            },\\
     \pi_{\s{up}} (\s{z}_1^{\ell})  
        &= \inf 
            \ppp{
                 \frac{c_{\s{I}}}{c_{\s{I}} + c_{\s{II}}} \le \pi \le 1 
                : \bar{s}_{\ell}\p{\pi, \s{z}_1^{\ell}} 
                    = c_{\s{I}} (1 - \pi)
            }.
    \end{align}
     Furthermore, we have,
    \begin{align}
    \label{eq:dinamic_programming}
        \bar{s}_{\ell} (\pi, \s{z}_1^{\ell}) 
            &= \min \left\{\vphantom{\E{
                        \bar{s}_{\ell+1}(\Pi_{\ell + 1} , \calF_{\ell+1})
                        \vert
                        \Pi_{\ell} = \pi,
                        \calF_{\ell} = \s{z}_1^{\ell}
                    }}
                    g(\pi) \:,c\pi \right.\nonumber\\
                    &\left.+\E{
                        \bar{s}_{\ell+1}(\Pi_{\ell + 1} , \calF_{\ell+1})
                        \vert
                        \Pi_{\ell} = \pi,
                        \calF_{\ell} = \s{z}_1^{\ell}
                    }\right\}.
    \end{align}
\end{theorem}

The equivalence of our problem to the well-known optimal stopping problem in Theorem \ref{theorem:OptimalStoppingProblem} allows us to achieve an optimal quickest solution. Theorem \ref{theorem:OptimalStoppingTime} states that our solution is represented by time dependent pairs of lower and upper thresholds $(\pi_{\s{low}}, \pi_{\s{up}})$, where each pair corresponds to a sequence of the observed edges $\s{z}_1^{\ell}$. The posterior probability thresholds in Theorem~\ref{theorem:OptimalStoppingTime} are calculated in real-time and are updated at every iteration $\ell$ according to the solution of the Bellman equation \eqref{eq:dinamic_programming}. The stopping rule in \eqref{eq:optimalStoppingTime} keeps track of the posterior probability $\Pi_{\ell}$. Once this posterior goes out of the threshold range, the auditing stops and makes a decision according to \eqref{eq:decision_rule}. 

\subsection{Statistical guarantees}

In this subsection we provide a few statistical guarantees associated with the optimal decision rule we derived in the previous subsection. We start by the following observation where we represent the optimal decision rule as a \textit{sequential probability
ratio test (SPRT)}, as SPRT’s are known to exhibit minimal expected stopping time among all sequential decision rules having given error probabilities.

\begin{theorem}[SPRT representation] \label{theorem:SPRT}
Consider the optimization problem in \eqref{eq:OptimalStoppingProblem}. If $\pi_{\s{low}}(\s{z}_1^{\ell}) < \pi < \pi_{\s{up}}(\s{z}_1^{\ell})$, for all $\ell = 1, 2, \ldots$, then the optimal solution given in Theorem~\ref{theorem:OptimalStoppingTime} can be equivalently defined as an SPRT with boundaries $\s{B}_{\s{low}}$ and $\s{B}_{\s{up}}$, namely,
\begin{align}
    \s{T}^\star = 
        \inf 
            \ppp{
                \ell\in\mathbb{N}: 
                    \Lambda_{\ell} \notin 
                    (\s{B}_{\s{low}}(\s{z}_1^{\ell}) , 
                    \s{B}_{\s{up}}(\s{z}_1^{\ell}))},
                    \label{eq:SPRT_stop_time}
\end{align}
where
\begin{align}
    \Lambda_{\ell} 
        \triangleq \prod_{i = 1}^{\ell} 
            \frac
                {\P{\s{Z}_{i} \vert \calF_{i - 1}, \calH_{1}}}
                {\P{\s{Z}_{i} \vert \calF_{i - 1}, \calH_{0}}} 
        = \frac
            {A_1(\s{Z}_{\ell} \vert \calF_{\ell-1})}
            {A_0(\s{Z}_{\ell} \vert \calF_{\ell-1})} 
            \Lambda_{\ell - 1},
\end{align}
for $\ell = 1, 2, \ldots$, with $\Lambda_0 \triangleq 1$. The thresholds $\s{B}_{\s{low}}$ and $\s{B}_{\s{up}}$ are given by
\begin{align}
    \s{B}_{\s{low}}(\s{z}_1^{\ell})
        &= \frac {1-\pi} {\pi}
            \cdot
            \frac
                { \pi_{\s{low}}(\s{z}_1^{\ell}) }
                { 1 - \pi_{\s{low}}(\s{z}_1^{\ell}) },
                \label{eq:Y_low}\\
    \s{B}_{\s{up}}(\s{z}_1^{\ell})
        &= \frac {1-\pi} {\pi}
            \cdot
            \frac
                { \pi_{\s{up}}(\s{z}_1^{\ell}) }
                { 1 - \pi_{\s{up}}(\s{z}_1^{\ell}) }.
                \label{eq:Y_up}
\end{align}
The decision rule is given by,
\begin{align}
    \delta_{\s{T}^\star}(\s{z}_1^{\s{T}^\star})
    =   \begin{cases}
            0,  &   \Lambda_{\s{T}^\star} \le \s{B}_{\s{low}}(\s{z}_1^{\s{T}^\star})\\
            1,  &   \Lambda_{\s{T}^\star} > \s{B}_{\s{up}}(\s{z}_1^{\s{T}^\star}).
            \label{eq:SPRT_decision_rule}
        \end{cases}
\end{align}
\end{theorem}

\begin{proposition}\label{prop:ErrorProb&Boundries}
    For each $\ell$, fix $0 < \s{B}_{\s{low}} \le 1 \le \s{B}_{\s{up}} < \infty$. Let,
    \begin{align}
        \s{P}_{e,1} &\triangleq \P_{\calH_0}{\delta_{\s{T}^\star} = 1},\\
        \s{P}_{e,2} &\triangleq \P_{\calH_1}{\delta_{\s{T}^\star} = 0}.
    \end{align}
    Then, the following relationship among the thresholds and the error probabilities hold,
    \begin{align}
        \s{B}_{\s{low}} &\ge \frac{\s{P}_{e,2}}{1 - \s{P}_{e,1}},\label{eq:LowerBoundIneq} \\ 
        \s{B}_{\s{up}} &\le \frac{1 - \s{P}_{e,2}}{\s{P}_{e,1}}. \label{eq:UpperBoundIneq}
    \end{align}
\end{proposition}
An important consequence of the above proposition is that the boundaries $\s{B}_{\s{low}}$ and $\s{B}_{\s{up}}$ can be chosen to yield a given level of error probability performance. For example, if we wish to design a test $\varphi$ with approximate error probabilities $p_{\varphi}$ and $q_{\varphi}$, then we can use Wald's approximations to choose boundaries $\s{B}_{\s{low}} = \frac{q_{\varphi}}{1 - p_{\varphi}}$ and $\s{B}_{\s{up}} = \frac{1 - q_{\varphi}}{p_{\varphi}}$. Then, inequalities \eqref{eq:LowerBoundIneq} and \eqref{eq:UpperBoundIneq} imply that the actual error probabilities are bounded according bu 
\begin{align}
    \s{P}_{e,1} 
        &\le \frac
                {p_{\varphi}}
                {1 - q_{\varphi}}
        = p_{\varphi} (1 + O(q_{\varphi})),\\
    \s{P}_{e,2} 
        &\le \frac
                {q_{\varphi}}
                {1 - p_{\varphi}}
        = q_{\varphi} (1 + O(p_{\varphi})).
\end{align}
Thus, for a small desired error probabilities, the actual error probabilities obtained by using Wald's approximations can be bounded by values that are quite close to their desired values. And, in fact, these bounds will be tight in the limit of small error probabilities. Namely, we can design a test that achieves with good accuracy error probabilities as small as desired. Clearly, as the error probabilities get smaller, the range $(\s{B}_{\s{low}},\s{B}_{\s{up}})$ gets larger so that the stopping time would increase. This is consistent with \eqref{eq:distingushing}, which shows that the hypothesis can be accurately distinguished by the limit value of the likelihood ratio.

\subsection{Model and data driven algorithms}
In this subsection, we present our model and data driven algorithms. The pseudo-codes of our misinformation detection procedure are given in Algorithms~\ref{Algo:MisinformationDetection_train} and \ref{Algo:ExactMisinformationDetection}. Specifically, Algorithm~\ref{Algo:MisinformationDetection_train} is an \emph{offline} procedure that trains an edge-classifier $f(\cdot)$, learns the transition probabilities $\alpha_{\calI}(\cdot \vert \cdot)$, and the initial probabilities $\eta_{\calI}(\cdot)$, for $\calI=\{0,1\}$. The required dataset contains both the social media graph and genuine/fake labeled information spreading traces $\{t_k\}_{k=1}^N$, that include for each user the followee from whom the information was retweeted. The \emph{online} Algorithm~\ref{Algo:ExactMisinformationDetection} is an implementation of our proposed sequential detection procedure. We next discuss the offline and online routines in more detail. 

\subsubsection{Offline Algorithm}
Our offline algorithm requires training dataset that contains both the social media graph $\calG$ and $N$ labeled information spreading traces, which we denote by $\{t_k\}_{k=1}^N$. The spreading traces are labeled as genuine or fake news and must include for each user the followee from whom the information was retweeted. The social media graph data includes all the connections between the users involved in the information spreading traces with each user has an associated feature vector. 
In order to find the function $f()$, that classifies an edge $e = (u,v)$ to one of $\s{Z}$ classes, we first train a linear SVM that classifies the information to genuine or fake news. We calculate the average feature vector of each information trace, and along with its corresponding label we generate the input for our SVM. The SVM returns a value in $[0,1]$. This range is divided into $\s{Z}$ equal intervals that are assigned with an edge type in a sequential manner. For example, if $\s{Z}=4$, then we use the mapping $[0,0.25] \to 0$, $(0.25, 0.5] \to 1$, $(0.5, 0.75] \to 2$, and $(0.75, 1] \to 3$. Using this mapping we then estimate two important terms:
\begin{itemize}
    \item \textbf{Initial probability:} the probability of an edge of type $\s{z}\in\calZ$ to forwarded information of type $\calI = \{0,1\}$ directly from the source $s$, using
    \begin{align}
    \label{eq:init_prob}
        \hat{\eta}_{\calI}(\s{z}) = \frac{\sum_{k=1}^N \sum_{\ell=1}^{|t_k|-1} \mathds{1}_{\{e_{\ell}^{t_k} \in \calE_{s^{t_k}}, W_{e_\ell^{t_k}}=\s{z}\}} \mathds{1}_{\{l_k=\calI\}} }{\sum_{k=1}^N |\calE_{s^{t_k}}|\mathds{1}_{\{l_k=\calI\}}},
    \end{align}
    where $e_{\ell}^{t_k}=(u_{\ell}^{t_k}, v_{\ell}^{t_k})$, and $\calE_{s^{t_k}}$ is the set of edges that are connected to the source $s^{t_k}$ in $t_k$.
    \item \textbf{Transition probabilities:} we estimate
    \begin{align} \label{eq:transition_prob}
        \hat{\alpha}_{\calI}(\s{z} \vert \s{z}') 
            = \frac
                {\sum_{k=1}^{N} \sum_{\ell=1}^{|t_k|-1} \mathds{1}_{\{W_{e_{\ell}^{t_k}}=\s{z}, W_{{e'_{\ell}}^{t_k}}=\s{z}'\}} \mathds{1}_{\{l_k=\calI\}}}
                {\sum_{k=1}^N\sum_{\ell=1}^{|t_{k}|-1} \mathds{1}_{\{{W}_{{e'}_{\ell}^{t_k}}=\s{z}'\}} \mathds{1}_{\{l_k=\calI\}}},
    \end{align}
    where ${e'}_{\ell}^{t_k}$ is the adjacent edge of $e_{\ell}^{t_k}$ that previously forwarded the information in the trace $t_k$; if no such edge exist, and $e_{\ell}^{t_k}$ is connected directly to the source $s^{t_k}$, then the indicator is nullified.
\end{itemize}

\subsubsection{Online Algorithm}
The inputs to our online procedure are: the social media graph $\calG$, partial information trace $\{\s{Z}_{\ell}\}_{\ell\ge 1}$ including the source $s$, edge classifier $f(\cdot)$, initial probabilities $\hat{\eta}_{\calI}(\cdot)$, and transition probabilities $\hat{\alpha}_{\calI}(\cdot \vert \cdot)$. The algorithm initializes the prior distribution of hypothesis $\calH_1$ according to the data, $\Pi_0 = \pi_0$. When an event $\s{Z}_{\ell}$ occurs, the algorithm calculates the conditional transition probability $\calA_{\calI}$ using \eqref{eq:NextObservationConditionalProbability}, and then updates $\Pi_{1}$ using \eqref{eq:SimpleFormRecursiveEquation}. We note that the exact calculation of \eqref{eq:dinamic_programming}, and therefore the thresholds $\pi_{\s{low}}, \pi_{\s{up}}$, entails taking expectation over the entire collection of unobserved edges in $\calE$. Therefore, our online algorithm performs a first-order approximation and stops with the first sign of convergence of $\Pi_{\ell}$, that is, it stops when $\abs{\Pi_{\ell+1}-\Pi_{\ell}}<\epsilon$, for some initialized $\epsilon>0$. The information is declared to be fake news if $\Pi_1 \ge \pi_1$, and genuine information, otherwise. Clearly, this approximation can only impair the performance of our algorithm.

\begin{algorithm}[!ht]
    \caption{Misinformation training (Offline)}
    \label{Algo:MisinformationDetection_train}
    \textbf{Input:} Social media graph, 
    $N$ information propagation traces $\{t_1, \ldots, t_N\}$ with labels $L = \{l_1, \ldots, l_N\}, l_k\in\{0,1\}$. Each $t_k$ is a sequence of $\abs{t_k}$ users $v_{\ell}^{t_k}$ and feature vectors $\mathbf{x}_{v_{\ell}^{t_k}}$.\\
    \textbf{Output:} $f(\cdot)$, $\hat{\alpha}_{\calI}(\cdot \vert \cdot)$,
    $\hat{\eta}_{\calI}(\cdot)$.
    \begin{algorithmic}
      \State $\bullet$ For each user $v_{\ell}$ in each information trace $t_k$, obtain the followee-follower feature vector $(\mathbf{x}_{u_{\ell}}, \mathbf{x}_{v_{\ell}})$ and compute $\hspace*{1.5cm}\bar{\mathbf{F}}^{(t_k)} = \frac{1}{|t_k|-1} 
                \begin{pmatrix}
                    \sum\limits_{\ell = 2}^{|t_k|} \mathbf{x}_{u_{\ell}^{t_k}}, & \sum\limits_{\ell = 2}^{|t_k|} \mathbf{x}_{u_{\ell}^{t_k}}
                \end{pmatrix}$.
        \State $\bullet$ Train edge classifier $f(\cdot)$ using SVM with input $(\bar{\mathbf{F}}, L)$.
        \State $\bullet$ Classify each edge $e=(u,v)\in\calE$ to $W_e \gets f(\mathbf{x}_u, \mathbf{x}_v)$.
        \State $\bullet$ For all $z \in \calZ$ calculate $\hat{\eta}_{\calI}(\s{z})$ using \eqref{eq:init_prob}.   
        \State $\bullet$ For all $z,z'\in \calZ$ calculate $\hat{\alpha}_{\calI}(\s{z} \vert \s{z}')$ using \eqref{eq:transition_prob}.
    \end{algorithmic}
\end{algorithm}
\begin{algorithm}[!ht]
    \caption{Misinformation Detection (Online)}
    \label{Algo:ExactMisinformationDetection}
    \textbf{Input:} Social media graph, partial trace $\{\s{Z}_{1},\s{Z}_2\ldots\}$ with a known source $s$, $f(\cdot)$, $\hat{\alpha}_{\calI}(\cdot \vert \cdot)$, $\hat{\eta}_{\calI}(\cdot)$.\\
    \textbf{Output:} \texttt{Genuine}/\texttt{Fake}.
    \begin{algorithmic}
        \State \textbf{Initialize:} $\epsilon$, $c_{\s{I}}$, $c_{\s{II}}$, $c$, $\Pi_{0} \gets \pi_1$, $\Pi_{1} \gets \pi_1 + 2\epsilon$, $\calF_0 \gets \emptyset$
        \While{$\abs{\Pi_0 - \Pi_1} \ge \epsilon $}
            \State $\Pi_0 \gets \Pi_1$, $\ell \gets \ell+1$, $\calF_{\ell} \gets \{\calF_{\ell-1}, \s{Z}_{\ell}\}$
            \State Calculate $\calA_{\calI}(\s{Z}_{\ell} \vert \calF_{\ell-1})$ using \eqref{eq:NextObservationConditionalProbability}.
            \State Update $\Pi_1$ using \eqref{eq:SimpleFormRecursiveEquation}. 
        \EndWhile
    \If {$\Pi_1 \ge \pi_1$}
    \Return \texttt{Fake}
    \Else
    \:\Return \texttt{Genuine}
    \EndIf
    \end{algorithmic}
\end{algorithm}

\section{Experiments} \label{section:Experiments}
In our research we make use of the \texttt{Weibo} dataset \cite{10.5555/3061053.3061153}. Sina Weibo is China's leading micro-blogging service provider with eight times more users than Twitter. The dataset includes 4,664 labeled information traces provided by Sina's community management center\footnote{https://service.account.\texttt{Weibo}.com} with an average of 816 retweets per trace. User features such as the number of followees, the number of followers, the registration days, etc., were originally extracted from Sina Weibo API.\footnote{4http://open.weibo.com/wiki/API} Table \ref{tab:Sina_Weibo} summarizes the details of the \texttt{Weibo} dataset. 
The social media graph structure $\calG$ is reconstructed by a union of all the information traces.
Moreover, recall that our online algorithm requires only partial information propagation traces in order to make a quick and accurate decision. Since the \texttt{Wiebo} dataset contains the entire information propagation trace, we have uniformly drawn 50\% of its observations. 

In our simulations, we divide the complete dataset into 80\% training and 20\% testing (with 10\% genuine news traces and 10\% fake news traces). In addition, we take $c_{\s{I}}= c_{\s{II}} = 10$, $c = 0.05$, $\epsilon= 0.001$, and $\s{Z}=4$. For simplicity of computation, in the calculation of $\calA_{\calI}$, we truncated the various paths in the graph to a predefined maximal length; note that this can only impair the performance of our algorithm.
\begin{table}[!t]
    \centering
    \begin{tabular}{|c| c|} 
    \hline
    \texttt{Number of users} &  2,746,818 \\
    \hline
    \texttt{Number of Tweets} &  3,805,656 \\
    \hline
    \texttt{Number of events} & 4,664 \\
    \hline
    \texttt{Number of rumors} & 2,313 \\
    \hline
    \texttt{Number of non-rumors} & 2,351 \\
    \hline
    \texttt{Average time length/event} &  2,460.7 Hours \\
    \hline
    \texttt{Average number of posts/event} &  816 \\
    \hline
    \texttt{Maximum number of posts/event} &  59,318 \\
    \hline
    \texttt{Minimum number of posts/event} & 10 \\
    \hline
\end{tabular}
\caption{Details of the \texttt{Weibo} dataset}
\label{tab:Sina_Weibo}
\end{table}
We compare our algorithm to QuickStop \cite{wei2019quickstop}, and the following state-of-the-art algorithms:
\begin{itemize}
    \item SVM-TS$_{u}$ and SVM-TS$_a$ \cite{ma2015detect} are dynamic series-time structure (DSTS) based SVM methods. The former takes in consideration the user features alone, while the later is a fully configured model that utilizes all content-based, user-based and diffusion-based features.
    \item DTC$_u$ and DTC$_a$ \cite{castillo2011information} are automatic methods for assessing the  credibility of a given set of tweets on Twitter based on decision trees. The first mentioned method uses the user features only, while the second method also uses the content-based features.
    \item SVM-RBF$_u$ and SVM-RBF$_a$ \cite{yang2012automatic} 
    are SVM-based detection methods with RBF kernel function. The former uses 
    only user features, whereas the later uses both user and content-based features.   
    \item CSI \cite{ruchansky2017csi} is a hybrid deep model which is composed of three modules: (1) an RNN to capture the temporal pattern of user activity, (2) fully connected layer for source characteristic learning based on the users behavior, and (3) integration module for classification.
    \item PPC-R, PPC-C and PPC-R+C \cite{liu2018early} are detection models through propagation path classification for time-series data with a gated recurrent unit (GRU), a CNN, and a combination of RNN and CNN, respectively.
\end{itemize}
For the purpose of comparison, we use the following performance metrics:
\begin{itemize}
  \item \emph{Accuracy:} the fraction of traces that are correctly classified.
  \item \emph{False positive (FP):} the fraction of genuine news classified as fake news.
  \item \emph{False negative (FN):} the fraction of fake news classified as genuine news.
  \item \emph{Detection time:} the average number of events required to declare the type of detection).
\end{itemize}

\begin{figure}[!t]
    \centering
    \begin{tikzpicture}
            \begin{axis}[
                xlabel={Number of events},
                ylabel={\textcolor{black}{$\pi_1$}},
                legend pos=north east,
                grid,
                width=0.8\textwidth,
                height=0.4\textwidth,
                enlarge x limits={rel=-0.5},
                y label style={at={(1.14,0.5)}, rotate=-90},
            ]
            \addplot[sharp plot, line width=0.5mm, mark=o, mark size=2pt, color=blue] coordinates {
                (0, 0.5)
                (1, 0.57911052649997)
                (2, 0.6543568976119262)
                (3, 0.7213864855900508)
                (4, 0.7839771747164417)
                (5, 0.4271416862175463)
                (6, 0.5031929707182154)
                (7, 0.5522557146017896)
                (8, 0.6230530258007558)
                (9, 0.6850170235143866)
                (10, 0.7529803903068902)
                (11, 0.7877788172619956)
                (12, 0.692322425203046)
                (13, 0.09830247074125889)
                (14, 0.13255164845596007)
                (15, 0.17639906915711343)
                (16, 0.04215003212962188)
                (17, 0.00243949798747322)
                (18, 0.0033060881511701758)
            
            };
            \addplot[sharp plot, line width=0.5mm, mark=triangle*, mark size=2pt, color=red] coordinates {
                (0, 0.5)
                (1, 0.8796690694959185)
                (2, 0.766147458417471)
                (3, 0.3226277144757841)
                (4, 0.7768803411661945)
                (5, 0.9621987260629985)
                (6, 0.9946547016351934)
                (7, 0.9881506494402302)
                (8, 0.9983623633791093)
                
                };
            \addplot[draw=black,  dashdotted] coordinates {(0, 0.5) (1, 0.5)
            (2, 0.5)
            (3, 0.5)
            (4, 0.5)
            (5, 0.5)
            (6, 0.5)
            (7, 0.5)
            (8, 0.5)
            (9, 0.5)
            (10, 0.5)
            (11, 0.5)
            (12, 0.5)
            (13, 0.5)
            (14, 0.5)
            (15, 0.5)
            (16, 0.5)
            (17, 0.5)
            (18, 0.5)
            };
            
            \legend{Real news, Fake news}
            \end{axis}
        \end{tikzpicture}
    \caption{
        Examples of $\Pi_{\ell}$ and $\s{T}$ on real/fake news under our method.}
    \label{fig:ExampAlgo}
\end{figure}
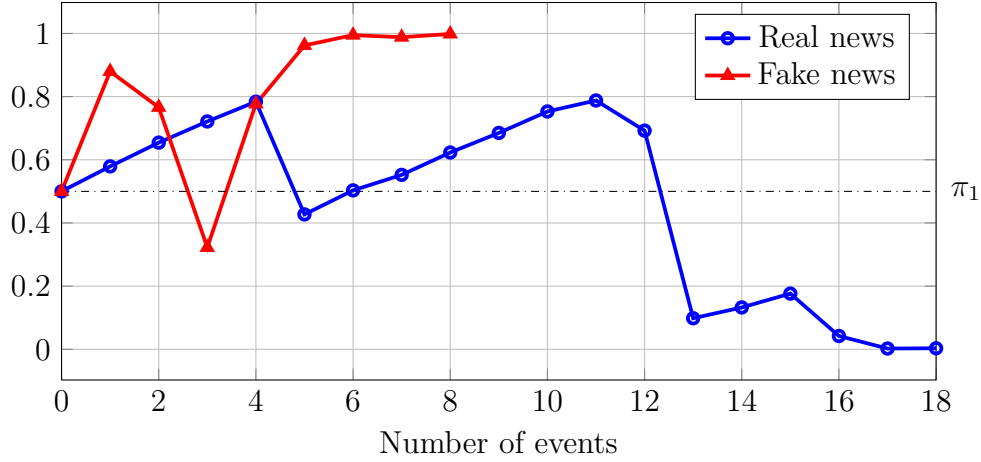

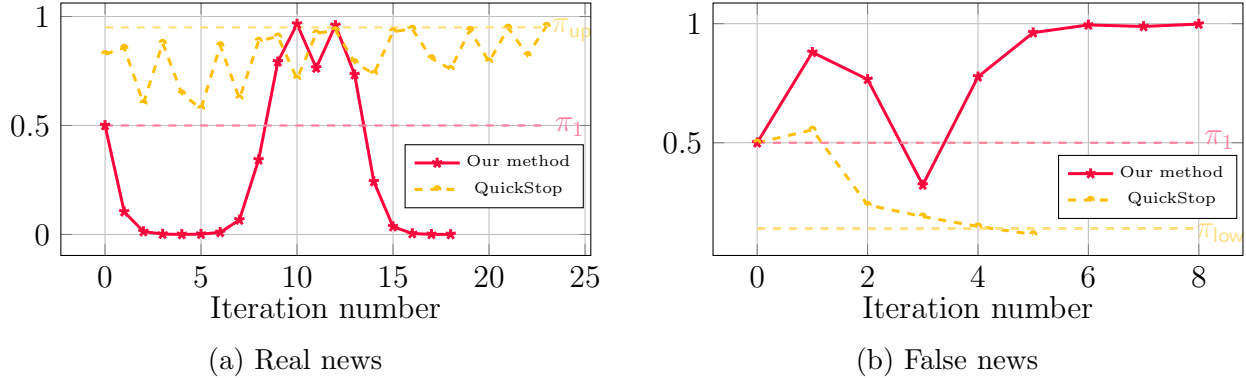
\begin{figure}[!t]
    \centering

    \begin{subfigure}{.475\linewidth} 
        \centering
        \begin{tikzpicture}
        \node at (6.8,3) {\textcolor{amber!50}{$\pi_{\s{up}}$}};
          \begin{axis}[
                xlabel={Iteration number},
                ylabel={\textcolor{americanrose!50}{$\pi_1$}},
                grid,
                width=1.1\textwidth,
                height=0.63\textwidth,
                legend style={font=\tiny, at={(axis cs:15.6,0.12)},anchor=south west},
                title style={at={(0.11,0.83)}},
                y label style={at={(1.135,0.51)}, rotate=-90},
                x label style={at={(0.5,0.04)}},
            ]
                \addplot[sharp plot, mark=star,  mark size=2pt, line width=0.4mm, color=americanrose] coordinates {
                (0, 0.5)
                (1, 0.10412717870219092)
                (2, 0.013329311183943333)
                (3, 0.001567732020715033)
                (4, 0.0001824700989797156)
                (5, 0.001332397988610235)
                (6, 0.009659168436683505)
                (7, 0.06655573624414732)
                (8, 0.34264226127582414)
                (9, 0.7921211084992742)
                (10, 0.9653455078332375)
                (11, 0.764024753547671)
                (12, 0.9594635732630571)
                (13, 0.7334090074917247)
                (14, 0.2422840976660464)
                (15, 0.035833424952802936)
                (16, 0.004301124414500019)
                (17, 0.0005018266640232949)
                (18, 5.8353125900759445e-05)
                };
                \addplot[sharp plot, mark=o, mark size=1pt, line width=0.4mm, dashed, color=amber] coordinates {
                    (0, 0.8273231622746187)
                    (1, 0.8560853971269164)
                    (2, 0.6010796905093045)
                    (3, 0.878333148368041)
                    (4, 0.6464692538617706)
                    (5, 0.5766612391784045)
                    (6, 0.8671343896256931)
                    (7, 0.6230869627822202)
                    (8, 0.887897744302166)
                    (9, 0.907695967214325)
                    (10, 0.7135395646461589)
                    (11, 0.9226857897372275)
                    (12, 0.9367776160377403)
                    (13, 0.789614503011769)
                    (14, 0.7365532746992783)
                    (15, 0.9305328840831372)
                    (16, 0.9432823630026582)
                    (17, 0.8081596721268453)
                    (18, 0.7583442685524602)
                    (19, 0.937637359182153)
                    (20, 0.7920311935120591)
                    (21, 0.948043137739061)
                    (22, 0.8221236271905716)
                    (23, 0.9567926273274936)
                    };
                \addplot[draw=americanrose!50, dashed, line width=0.3mm] coordinates {(0, 0.5) (1, 0.5)
                (2, 0.5)
                (3, 0.5)
                (4, 0.5)
                (5, 0.5)
                (6, 0.5)
                (7, 0.5)
                (8, 0.5)
                (9, 0.5)
                (10, 0.5)
                (11, 0.5)
                (12, 0.5)
                (13, 0.5)
                (14, 0.5)
                (15, 0.5)
                (16, 0.5)
                (17, 0.5)
                (18, 0.5)
                (19, 0.5)
                (20, 0.5)
                (21, 0.5)
                (22, 0.5)
                (23, 0.5)
                };
                \addplot[draw=amber!50, dashed, line width=0.3mm] coordinates {
                (0, 0.95) 
                (1, 0.95)
                (2, 0.95)
                (3, 0.95)
                (4, 0.95)
                (5, 0.95)
                (6, 0.95)
                (7, 0.95)
                (8, 0.95)
                (9, 0.95)
                (10, 0.95)
                (11, 0.95)
                (12, 0.95)
                (13, 0.95)
                (14, 0.95)
                (15, 0.95)
                (16, 0.95)
                (17, 0.95)
                (18, 0.95)
                (19, 0.95)
                (20, 0.95)
                (21, 0.95)
                (22, 0.95)
                (23, 0.95)
            };
                \legend{Our method, QuickStop}
            \end{axis}
        \end{tikzpicture}
        \caption{Real news}
        \label{fig:Pi_evo_comp_news}
    \end{subfigure}\hfill
    \begin{subfigure}{.475\linewidth} 
        \centering
        \begin{tikzpicture}
        \node at (6.75,0.3) {\textcolor{amber!50}{$\pi_{\s{low}}$}};
            \begin{axis} [
                xlabel={Iteration number},
                ylabel={\textcolor{americanrose!50}{$\pi_1$}},
                grid,
                width=1.1\textwidth,
                height=0.63\textwidth,
                legend style={font=\tiny, at={(axis cs:5.45,0.18)}, anchor=south west},
                title style={at={(0.11,0.83)}},
                y label style={at={(1.13,0.46)}, rotate=-90},
                x label style={at={(0.5,0.04)}},
            ]
            \addplot[sharp plot, mark=star, mark size=2pt, line width=0.4mm, color=americanrose] coordinates {
            (0, 0.5)
            (1, 0.8796690694959185)
            (2, 0.766147458417471)
            (3, 0.3226277144757841)
            (4, 0.7768803411661945)
            (5, 0.9621987260629985)
            (6, 0.9946547016351934)
            (7, 0.9881506494402302)
            (8, 0.9983623633791093)
            };
            \addplot[sharp plot, mark=o, dashed,  mark size=1pt, line width=0.4mm, color=amber] coordinates {
            (0, 0.5)
            (1, 0.5538840617299962)
            (2, 0.23924783113174808)
            (3, 0.18980456164149387)
            (4, 0.14858360837437565)
            (5, 0.11504368132169947)
            };
            \addplot[draw=americanrose!50, dashed, line width=0.3mm] coordinates {(0, 0.5) (1, 0.5)
            (2, 0.5)
            (3, 0.5)
            (4, 0.5)
            (5, 0.5)
            (6, 0.5)
            (7, 0.5)
            (8, 0.5)
            };
            \addplot[draw=amber!50, dashed, line width=0.4mm] coordinates {
            (0, 0.14) 
            (1, 0.14)
            (2, 0.14)
            (3, 0.14)
            (4, 0.14)
            (5, 0.14)
            (6, 0.14)
            (7, 0.14)
            (8, 0.14)
            };  
            \legend{Our method, QuickStop}
            \end{axis}
        \end{tikzpicture}
        \caption{False news }
        \label{fig:Pi_evo_comp_fake}
    \end{subfigure}
    
    \caption{Examples of $\Pi_{\ell}$ and stopping time $\s{T}$ under our method and QuickStop.}
    \label{fig:Pi_evo_comp}
\end{figure}

Figures~\ref{fig:ExampAlgo} and \ref{fig:Pi_evo_comp} present the evolution of $\Pi_{\ell}$ on two traces, genuine and fake news, chosen from the \texttt{Weibo} dataset, using our method, and as compared to QuickStop, respectively. Both examples evidently show that our algorithm succeeds and declares the correct information type, while QuickStop fails to do so, as it yields FN (see, Fig.~\ref{fig:Pi_evo_comp_news}) and FP (see, Fig.~\ref{fig:Pi_evo_comp_fake}).

As expected, it is evident that our algorithm makes a quicker decision when a fake news propagates, due to the propagation cost of spreading misinformation in \eqref{eq:OptimalStoppingProblem}. Specifically, based on our experiments, our detection algorithm requires an average of 5.6 events for misinformation detection and an average of 7.2 events for news detection; altogether, a decision is made after 6.29 events on average. Keeping in mind that in our simulations, we implemented a first-order approximation of the exact algorithm in Theorem \ref{theorem:OptimalStoppingTime}, it is reasonable that the likelihood $\Pi_{\ell}$ may exceed the interval $(\pi_{\s{low}} (\s{z}_1^{\ell}), \pi_{\s{up}} (\s{z}_1^{\ell}))$, before the 
convergence of $\Pi_{\ell}$ occurs. In this case, our exact algorithm will arrive at quicker decisions with the same accuracy rate.

Finally, Fig.~\ref{fig:ten_algorithms} compares the accuracy of our algorithm to the previously mentioned algorithms. It is clear that our algorithm outperforms all of these algorithms, both in terms of accuracy and detection time.
Specifically, in Table~\ref{tab:us_vs_QuickStop} we zoom-in and compare our algorithm to QuickStop. It can be seen that on average, our algorithm achieves the same detection accuracy (as well as false positive (FP) and false negative (FN) rates), but roughly in half the time.

\begin{figure}[!t]
    \centering
    \begin{tikzpicture}
        \begin{axis}[
            xlabel=Number of Events,
            ylabel=Accuracy,
            legend pos=north west,
            grid,
            width=0.5\textwidth,
            height=0.5\textwidth,
            legend style={at={(1.1,0.822)}, anchor=north west, font=\footnotesize},
        ]
        
        \addplot[sharp plot, mark=o, mark size=2pt, line width=0.2mm, color=blue, dashed] coordinates {
            (50, 0.53)
            (100, 0.60)
            (150, 0.62)
            (200, 0.64)
            (250, 0.7)
            (300, 0.67)
            (350, 0.66)
            (400, 0.7)
            (450, 0.66)
            (500, 0.72)
        };
        
        \addplot[sharp plot, mark=o, mark size=2pt, line width=0.2mm, color=blue] coordinates {
            (50, 0.56)
            (100, 0.56)
            (150, 0.68)
            (200, 0.66)
            (250, 0.69)
            (300, 0.75)
            (350, 0.65)
            (400, 0.70)
            (450, 0.76)
            (500, 0.71)
        };
        
        \addplot[sharp plot, mark=triangle, mark size=2pt, line width=0.2mm, color=red, dashed] coordinates {
            (17, 0.64)
            (50, 0.62)
            (100, 0.71)
            (150, 0.705)
            (200, 0.73)
            (250, 0.75)
            (300, 0.74)
            (350, 0.78)
            (400, 0.77)
            (450, 0.81)
            (500, 0.80)
        };
        
        \addplot[sharp plot, mark=triangle, mark size=2pt, line width=0.2mm,  color=red] coordinates {
            (17, 0.595)
            (50, 0.68)
            (100, 0.65)
            (150, 0.75)
            (200, 0.73)
            (250, 0.725)
            (300, 0.80)
            (350, 0.795)
            (400, 0.79)
            (450, 0.82)
            (500, 0.825)
        };
        
        \addplot[sharp plot, mark=diamond, mark size=2pt, color=yellow, line width=0.2mm, dashed] coordinates {
            (17, 0.495)
            (50, 0.635)
            (100, 0.77)
            (150, 0.78)
            (200, 0.79)
            (250, 0.803)
            (300, 0.82)
            (350, 0.84)
            (400, 0.845)
            (450, 0.82)
            (500, 0.855)
        };
        
        \addplot[sharp plot, mark=diamond, mark size=2pt, line width=0.2mm, color=yellow] coordinates {
            (17, 0.49)
            (50, 0.545)
            (100, 0.79)
            (150, 0.81)
            (200, 0.83)
            (250, 0.84)
            (300, 0.845)
            (350, 0.86)
            (400, 0.86)
            (450, 0.86)
            (500, 0.865)
        };

        \addplot[sharp plot, mark=square, mark size=2pt, line width=0.2mm, color=black] coordinates {
            (17, 0.79)
            (50, 0.825)
            (100, 0.875)
            (150, 0.85)
            (200, 0.91)
            (250, 0.905)
            (300, 0.91)
            (350, 0.915)
            (400, 0.92)
            (450, 0.93)
            (500, 0.91)
        };

        \addplot[sharp plot, mark=oplus, mark size=2pt, line width=0.2mm,  color=green] coordinates {
            (17, 0.615)
            (50, 0.835)
            (100, 0.855)
            (150, 0.865)
            (200, 0.867)
            (250, 0.87)
            (300, 0.867)
            (350, 0.867)
            (400, 0.866)
            (450, 0.865)
            (500, 0.86)
        };

        \addplot[sharp plot, mark=oplus, mark size=2pt, line width=0.2mm,  color=green, dashed] coordinates {
            (17, 0.71)
            (50, 0.78)
            (100, 0.80)
            (150, 0.83)
            (200, 0.84)
            (250, 0.845)
            (300, 0.84)
            (350, 0.845)
            (400, 0.835)
            (450, 0.83)
            (500, 0.825)
        };

        \addplot[sharp plot, mark=+, mark size=2pt, line width=0.2mm, color=cyan] coordinates {
            (17, 0.69)
            (50, 0.79)
            (100, 0.815)
            (150, 0.83)
            (200, 0.85)
            (250, 0.86)
            (300, 0.855)
            (350, 0.854)
            (400, 0.845)
            (450, 0.845)
            (500, 0.85)
        };
        
        \addplot[sharp plot, mark=*, mark size=2pt, line width=0.2mm, color=amber] coordinates {
            (12.75, 0.85)
        };
        
        \addplot[sharp plot, mark=star, mark size=4pt, line width=0.2mm, color=americanrose] coordinates {
            (6.29, 0.86)
        };
        
        \legend{SVM-TS$_u$, SVM-TS$_a$, DTC$_u$, DTC$_a$, SVM-RBF$_u$, SVM-RBF$_a$, CSI, PPC-R, PPC-C, PPC-R+C,  Quickstop, Our method}
        \end{axis}
    \end{tikzpicture}
    \caption{Detection accuracy as a function of $\ell$.}
    \label{fig:ten_algorithms}
\end{figure}
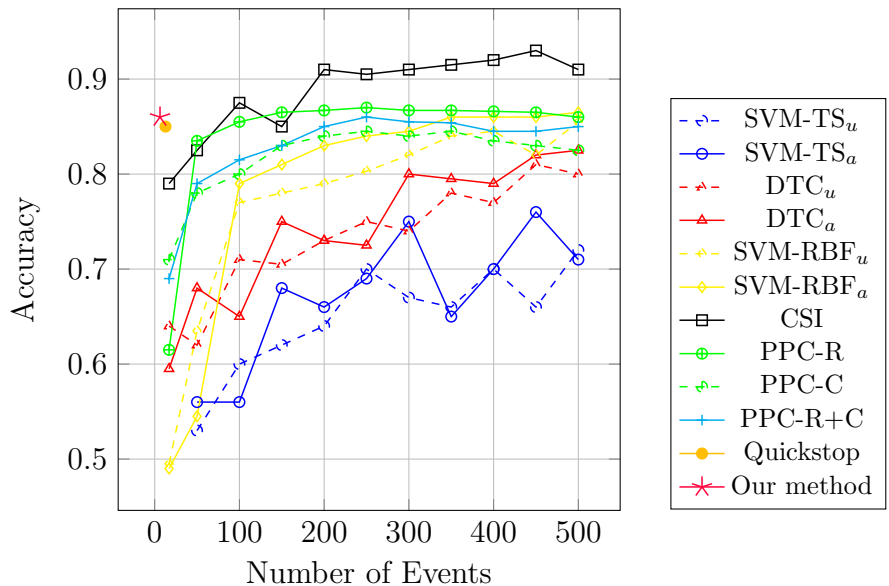
\begin{table}[t!]
    \centering
    \begin{tabular}{|c|c|c|c|c|}
        \hline
        Method & Accuracy & FP & FN & Decision Deadline \\
        \hline
        \hline
        Quickstop & 0.85 & 0.08 & 0.20 & 12.75 \\
        \hline
        \textbf{Ours} & \textbf{0.86} & \textbf{0.08} & \textbf{0.18} & \textbf{6.29} \\
        \hline
    \end{tabular}
    \caption{Detailed comparison of our method and QuickStop.}
    \label{tab:us_vs_QuickStop}
\end{table}


\section{Proofs}

\subsection{Proof of Theorem~\ref{thm:recursive}}
Recall that $\Pi_{\ell}=\P{\calH_1 \vert \calF_{\ell}}$, and note that $\Pi_{\ell}$ is a Doob's martingale with respect to the filtration $\calF_{\ell}$. Indeed,
\begin{align}
    \E{\Pi_{\ell} \vert \calF_{\ell-1}} 
        &=  \E{\E{\mathds{1}_{{\calH_{1}}} \vert \calF_{\ell}} \vert \calF_{\ell-1}} 
        =   \E{\mathds{1}_{{\calH_{1}}} \vert \calF_{\ell-1}} 
        =   \Pi_{\ell-1}, \label{eq:PiMartingaleProof}
\end{align}
where we used the fact that $\calF_{\ell-1}\subset\calF_{\ell}$. According to the Bayes rule, we have

\begin{align}
    \Pi_{\ell} 
        &=  \P{\calH_1 \vert \calF_{\ell}}\\
        &=  \frac
                {\P{\calF_{\ell} \vert \calH_1} \P{\calH_1}}
                {\P{\calF_{\ell}}}\\
        &=  \frac
                {\pi_1 \P{\calF_{\ell} \vert \calH_1}}
                {\pi_1 \P{\calF_{\ell} \vert \calH_1} + \pi_0 \P{\calF_{\ell} \vert \calH_0}}, \label{Eq:Hypothesis1Probability}
\end{align}
while the above joint probability distributions can be expressed as,
\begin{align}
    \P{\calF_{\ell} \vert \calH_{\calI}} 
        &=  \P{\s{Z}_1 , \s{Z}_2 , \ldots , \s{Z}_{\ell} \vert \calH_{\calI}} \\ 
        &=  \prod\limits_{i = 1}^{\ell} 
                A_{\calI}(\s{Z}_{i} \vert \calF_{i-1}).\label{Eq:JointProb}
\end{align}
Thus, we can write
\begin{align}
    \frac{1 - \Pi_{\ell}}{\Pi_{\ell}} 
        &= \frac{
                (1-\pi_1)      
                \prod\limits_{i = 1}^{\ell}
                    A_{0}(\s{Z}_{i} \vert \calF_{i - 1})
                } 
                {\pi_1 \prod \limits_{i = 1}^{\ell} A_{1}(\s{Z}_{i} \vert \calF_{i - 1})},
\end{align}
which implies that
\begin{align}
    \frac{1 - \Pi_{\ell + 1}}{\Pi_{\ell + 1}} 
        &= \frac{1 - \Pi_{\ell}}{\Pi_{\ell}} \frac{A_0(\s{Z}_{\ell + 1} \vert \calF_{\ell})}{A_1(\s{Z}_{\ell + 1} \vert \calF_{\ell})}.
\end{align}
Therefore, we arrive at the following recursive relation,
\begin{align}
    &\Pi_{\ell + 1} 
        = \frac{\Pi_{\ell} 
            A_1(\s{Z}_{\ell + 1} \vert \calF_{\ell})}{\Pi_{\ell} 
            A_1(\s{Z}_{\ell + 1} \vert \calF_{\ell}) 
            + (1 - \Pi_{\ell})
            A_0(\s{Z}_{\ell+1} \vert \calF_{\ell})}. \label{eq:SimpleFormRecursiveEquationproof}
\end{align}
It is left to derive an explicit formula for $A_{\calI}(\s{Z}_{i} \vert \calF_{i-1})$. Recall that $\calP_{\ell}$ is the set of all the possible directed paths starting from the source $s\in\calV$ and ending at the current observation $\s{Z}_{\ell}$. Then, using the law of total probability, $A_{\calI}(\s{Z}_{\ell} \vert \calF_{\ell-1})$ can be explicitly rewritten as,  
\begin{align} \label{eq:A_Idefinition}
    A_{\calI}(\s{Z}_{\ell} \vert \calF_{\ell-1}) 
        &=     \P{\s{Z}_{\ell} \vert \calF_{\ell-1} , \calH_{\calI}} \\
        &=             \sum\limits_{\s{P} \in \calP_{\ell}} 
                            \P{\s{P}\vert \calF_{\ell -1}, \calH_{\calI}}\P{\s{Z}_{\ell} \vert \calF_{\ell -1},\s{P}, \calH_{\calI}}.\label{eqn:summProb}
\end{align}
We next find formulas for the probabilities inside the summation in \eqref{eqn:summProb}, starting with $\P{\s{P}\vert \calF_{\ell -1}, \calH_{\calI}}$. There are two cases to consider here. If $\s{Z}_{\ell}$ is connected to the source $s$ directly, then we clearly have,
\begin{align}
    A_{\calI}(\s{Z}_{\ell} \vert \calF_{\ell-1})= \eta_\calI(\s{Z}_{\ell}).\label{eqn:directConc}
\end{align}
Otherwise, let us denote the set of observations in the sequence $(\s{Z}_1, \ldots, \s{Z}_{\ell-1})$ that belong to the path $\s{P}$ by $\calF_{\ell-1}^{\s{P}}$ and their indices by $\mathcal{J}_{\ell-1}^{\s{P}}$. Then the independency of the observations along with Bayes theorem and the law of total probability imply that,
\begin{align}
    &\P{\s{P}\vert \calF_{\ell -1}, \calH_{\calI}}
    = 
    \P{\s{P} \vert \calF^{\s{P}}_{\ell -1}, \calH_{\calI}}\\
    &\hspace{1cm}=
    \frac{ \prod_{i \in \mathcal{J}_{\ell-1}^{\s{P}}}     \P\p{\s{Z}_i \vert \calF_{i-1}, \s{P}, \calH_{\calI}}}
    {\sum_{\s{P}' \in \calP_{\ell}} \prod_{i \in \mathcal{J}_{\ell-1}^{\s{P}}} \P\p{\s{Z}_i \vert \calF_{i-1}, \s{P}', \calH_{\calI}}},
\end{align}

where we have used the fact that $\P(\s{P}\vert\calH_{\calI}) = |\calP_\ell|^{-1}$. Thus, we see that finding a formula for \eqref{eqn:summProb} boils down to finding a formula for $\P{\s{Z}_{i} \vert \calF_{i-1},\s{P}, \calH_{\calI}}$, for any $1\leq i\leq \ell$, and $\s{P}\in\calP_\ell$. By the Markov property, the conditional probability of $\s{Z}_{i}$ given the path $\s{P}$ and the sequence of observations $\calF_{i-1}$ depends only on the last observation in the sequence $(\s{Z}_1, \ldots, \s{Z}_{i-1})$ in the path $\s{P}$. Recall that we denote the index of this last observation by $\s{I}_{i}(\s{P})$. Also, recall that the sequence of observations $(\s{Z}^{\s{P}}_{\s{I}_{i}(\s{P})},\s{W}^{\s{P}}_{\s{I}_{i}(\s{P})+1\to i-1},\s{Z}^{\s{P}}_{i})$ across the path $\s{P}$ form a first-order Markov chain. Otherwise, by the law of total probability, we have
\begin{align}
    &\P{\s{Z}_{i} \vert \calF_{i - 1},\s{P}, \calH_{\calI}}\nonumber\\
    &\hspace{1cm}= \sum_{\s{z}_{\s{I}_{i}(\s{P})+1}^{i-1}\in\calZ^{i-\s{I}_{i}(\s{P})-1}}
    \prod_{j=\s{I}_{i}(\s{P})+1}^{i} \alpha_{\calI} (\s{z}_{j}\vert\s{z}_{j-1}),\label{eqn:posteriorEll}
\end{align}
where $\s{z}_{\s{I}_{i}(\s{P})+1}^{i-1}=(\s{z}_{\s{I}_{i}(\s{P})+1},\ldots,\s{z}_{i-1})$. 
Thus, we get \eqref{eqn:conditionalPath}, shown at the top of the next page.
\begin{align}
    \P{\s{P}\vert \calF_{\ell -1}, \calH_{\calI}} = \frac{\prod_{i\in \mathcal{J}_{\ell-1}^{\s{P}}} \sum_{\s{z}_{\s{I}_{i}(\s{P})+1}^{i-1}\in\calZ^{i-\s{I}_{i}(\s{P})-1}}
    \prod_{j=\s{I}_{i}(\s{P})+1}^{i} \alpha_{\calI} (\s{z}_{j}\vert\s{z}_{j-1})}{\sum_{\s{P}'\in\calP_{\ell}} \prod_{i\in\mathcal{J}_{\ell-1}^{\s{P}}}
    \sum_{\s{z}_{\s{I}_{i}(\s{P}')+1}^{i-1}\in\calZ^{i-\s{I}_{i}(\s{P}')-1}}
    \prod_{j=\s{I}_{i}(\s{P}')+1}^{i} \alpha_{\calI} (\s{z}_{j}\vert\s{z}_{j-1})}.\label{eqn:conditionalPath}
\end{align}
Substituting \eqref{eqn:directConc}, \eqref{eqn:posteriorEll}, and \eqref{eqn:conditionalPath} in \eqref{eqn:summProb}, we readily obtain an expression for $A_{\calI}(\s{Z}_{\ell} \vert \calF_{\ell-1})$. Specifically, recall the definition of the measure $\mu_{\calI}(\s{P})$ in \eqref{eqn:muDef}. Then,
\begin{align}
    &A_{\calI}(\s{Z}_{\ell} \vert \calF_{\ell-1})\nonumber\\
    &\hspace{0.5cm}= \E_{\mu_{\calI}}\pp{\sum_{\s{z}_{\s{I}_{i}(\s{P})+1}^{i-1}\in\calZ^{i-\s{I}_{i}(\s{P})-1}}
    \prod_{j=\s{I}_{i}(\s{P})+1}^{i} \alpha_{\calI} (\s{z}_{j}\vert\s{z}_{j-1})}.
\end{align}

\subsection{Proof of Theorem~\ref{theorem:OptimalStoppingProblem}}
We separate this proof into two parts, starting with the propagation cost. We first show that $\E{\s{T} \mathds{1}_{{\calH_1}}} = \E{\s{T} \Pi_{\s{T}}}$. According to the law of total expectation we have
\begin{align}
    \E{\s{T} \mathds{1}_{{\calH_1}}} 
        &= \E{
                \E{
                    \s{T} 
                    \mathds{1}_{{\calH_1}} 
                    \vert \s{T}
                }
            }\\
        &= \E{
                \s{T}
                \E{
                    \mathds{1}_{{\calH_1}}
                    \vert \s{T}
                }
            }\\
        &= \E{
                \s{T} 
                \E{
                    \E{ 
                        \mathds{1}_{{\calH_1}} 
                        \vert \calF_{\s{T}} , \s{T}
                    }
                }
            }\\
        &= \E{
                \s{T} 
                \E{
                    \E{ 
                        \mathds{1}_{{\calH_1}} 
                        \vert \calF_{\s{T}}
                    } 
                    \vert \s{T}
                }
            }\\
        &= \E{
                \s{T} 
                \E{
                    \P {\calH_1 \vert \calF_{\s{T}}} 
                    \vert \s{T}
                }
            }\\
        &= \E{ 
                \s{T} 
                \E{\Pi_{\s{T}} \vert \s{T}}
            }\\
        &= \E{\s{T} \Pi_{\s{T}}}.
\end{align}
Next, we analyze the average costs of errors. Specifically, we will show that $c_e (\s{T}, \delta_{\s{T}}) = \E{ \min\{c_{\s{II}} \Pi_{\s{T}} , c_{\s{I}} (1 - \Pi_{\s{T}})\}}$, for the likelihood ratio test $\delta_{\s{T}}$. Indeed,
\begin{align}
    &c_e (\s{T}, \delta_{\s{T}})
        =  c_{\s{I}}
            \P_{\calH_0}{\delta_{\s{T}} = 1} 
            + c_{\s{II}} 
            \P_{\calH_1}{\delta_{\s{T}} = 0}\\ 
        &= \sum\limits_{\ell = 1}^{\infty}   
                c_{\s{I}} 
                \P(\delta_{\ell} = 1 , \calH_0 \vert \s{T} = \ell)\P{\s{T} = \ell}\nonumber\\ 
                &\hspace{1cm}+ c_{\s{II}} 
                \P(\delta_{\ell} = 0 , \calH_1 \vert \s{T} = \ell)\P{\s{T} = \ell}\\
        &= \sum\limits_{\ell = 1}^{\infty}
                c_{\s{I}} 
                \E{ 
                    \mathds{1}_{\{\delta_{\ell} = 1\}}
                    \mathds{1}_{{\calH_0}}
                    \vert \s{T} = \ell
                }\P{\s{T} = \ell}\nonumber\\ 
                &\hspace{1cm}+ c_{\s{II}}
                \E{
                    \mathds{1}_{\{\delta_{\ell} = 0\}}
                    \mathds{1}_{{\calH_1}} 
                    \vert \s{T} = \ell
                }\P{\s{T} = \ell}
            \\
        &= \sum\limits_{\ell = 1}^{\infty} 
            (
                c_{\s{I}} 
                \E{
                    \E{
                        \mathds{1}_{\{ \delta_{\ell} = 1\}}
                        \mathds{1}_{{\calH_0}} 
                        \vert \calF_{\ell}
                    } 
                    \vert \s{T} = \ell
                } \nonumber\\
                &\qquad +   
                c_{\s{II}}
                \E{
                    \E{
                        \mathds{1}_{\{\delta_{\ell} = 0\}}
                        \mathds{1}_{{\calH_1}} 
                        \vert \calF_{\ell}
                    } 
                    \vert \s{T} = \ell
                } 
            )
            \P{\s{T} = \ell}\\
        &= \sum\limits_{\ell = 1}^{\infty}
                \E{ 
                    c_{\s{I}}
                    \mathds{1}_{\{\delta_{\ell}(\calF_{\ell}) = 1\}}
                    (1-\Pi_{\ell}) 
                    + c_{\s{II}}
                    \mathds{1}_{\{\delta_{\ell}(\calF_{\ell}) = 0\}}
                    \Pi_{\ell}
                    \vert \s{T} = \ell
                }\nonumber\\
                &\hspace{2cm}\cdot\P{\s{T} = \ell} \\
    &\ge \sum\limits_{\ell = 1}^{\infty}
            \E{ 
                \min{
                    \{  c_{\s{I}} (1 - \Pi_{\ell}) ,
                        c_{\s{II}} \Pi_{\ell}\}
                } 
                \vert \s{T} = \ell
            }
            \P{\s{T} = \ell} \label{eq:LiklihoodRatioEquality}\\ 
    &= \E{
            \min{
                \{  c_{\s{I}} (1 - \Pi_{\ell}) ,
                    c_{\s{II}} \Pi_{\ell}\}
            }
        }.
\end{align}
Finally, we note that \eqref{eq:LiklihoodRatioEquality} holds with equality for the likelihood-ratio test rule, i.e., $\delta_{\s{T}} = \mathds{1}_{\ppp{c_{\s{II}} \Pi_{\s{T}} > c_{\s{I}} (1 - \Pi_{\s{T}})}}$, which concludes the proof.

\subsection{Proof of Theorem~\ref{theorem:OptimalStoppingTime}}
Consider a two-dimensional stochastic process $\{\mathbf{X}_{\ell}:\ell = 0, 1, \ldots\}$ with state space $\mathbb{S} = [0, 1] \times \mathbb{Z}$. Let $\mathbf{X}_0 = [\pi,m]^T$, and consider a family of measures $\ppp{\P_{(\pi,m)}: [\pi,m]^T \in \mathbb{S}}$, such that 
\begin{align}
    \P_{(\pi,m)}{\mathbf{X}_0 = 
        [\pi,m]^T} 
    = 1.
\end{align}
We denote the expectation under $\P_{(\pi, m)}$ by $\E_{(\pi, m)}$. Note that the first entry of $\mathbf{X}_{\ell}$ evolves in time $\ell$ according to the recursion rule \eqref{eq:SimpleFormRecursiveEquation}, and the second entry increases by $\ell$ units, more specifically $\mathbf{X}_{\ell} = [\Pi_{\ell},m + \ell]^T$. Therefore, the problem in \eqref{eq:equivallenProb} is a special case of the following optimal stopping problem (with $m = 0$),
\begin{align}
    \sup\limits_{\s{T} \in \calT_1} 
        \E_{(\pi,m)}h(\mathbf{X}_{\s{T}}),
\end{align}
where 
\begin{align}
    h(\mathbf{x}) 
        \triangleq - g(\pi) - c m \pi,
            \quad \mathbf{x} = [x,\pi]^T \in \mathbb{S},
\end{align}
and recall that $g(\pi) = \min \{c_{\s{II}} \pi, c_{\s{I}}(1 - \pi)\}$. Since $\sup_{\s{T} \in \calT} h(\mathbf{X}_{\s{T}}) \le c(-m)^+ \pi$, the general infinite horizon case of the optimal stopping theory \cite{poor2008quickest} states that the stopping time that solves problem \eqref{eq:equivallenProb} is given by
\begin{align}
    \s{T}^\star = \inf \{\ell\in\mathbb{N}: h(\mathbf{X}_{\ell}) = \gamma_{\ell}(\mathbf{X}_{\ell}, \calF_{\ell})\},
\end{align}
where for $\ell=1,2,\ldots$,
\begin{align}
    \gamma_{\ell}(\mathbf{x}, \s{z}_1^{\ell}) 
        \triangleq  \sup \limits_{\s{T} \in \calT_{\ell}} 
            \E_{(\pi, m) \vert \calF_{\ell}}{ h(\mathbf{X}_{\s{T}}) 
                \vert 
                \calF_{\ell} = \s{z}_1^{\ell}
            }.
\end{align}
Clearly, we have $\gamma_{\ell} (\mathbf{x}, \s{z}_1^{\ell}) = - s_{\ell}(\pi, \s{z}_1^{\ell}) - c m \pi$, where
\begin{align}
    s_{\ell} (\pi, \s{z}_1^{\ell}) 
        &= - \sup\limits_{\s{T} \in \calT_{\ell}}  
            \E_{(\pi, 0) \vert \calF_{\ell}}{ h(\mathbf{X}_{\s{T}}) 
                \vert 
                \calF_{\ell} = \s{z}_1^{\ell}
            }\\
        &= \inf\limits_{\s{T} \in \calT_{\ell}}  
            \E{ g(\Pi_{\s{T}}) + c\s{T}\Pi_{\s{T}} 
                \vert 
                \Pi_{\ell} = \pi,
                \calF_{\ell} = \s{z}_1^{\ell}
            }.
\end{align}
Note that $s_{\ell}(\pi, \s{z}_1^{\ell})$ is the minimum expected total cost if the algorithm is obligated to stop at time $\s{T} \ge \ell$, conditioned on the information up to time $\ell$. According to the general infinite horizon case of the optimal stopping theory the sequence $\{\gamma_{\ell}\}$ satisfies the condition
\begin{align}
    \gamma_{\ell}(\mathbf{x}, \s{z}_1^{\ell}) 
        &= \max \left\{
                h(\mathbf{x}),\right.\nonumber\\
                &\left.\quad\E_{(\pi, m) \vert \calF_{\ell}}{
                    \gamma_{\ell+1}(\mathbf{X}_{\ell + 1} , \calF_{\ell+1})
                    \vert
                    \calF_{\ell} = \s{z}_1^{\ell}
                }\right\}.
\end{align}
Therefore, 
\begin{align}
    s_{\ell}(\pi, \s{z}_1^{\ell}) 
        &= \min \left\{
                g(\pi) + c \ell \pi,\right.\nonumber\\ 
                &\left.\E{
                    s_{\ell + 1}(\Pi_{\ell + 1} , \calF_{\ell+1})
                    \vert
                    \Pi_{\ell} = \pi,
                    \calF_{\ell} = \s{z}_1^{\ell}
                }\right\}.    
\end{align}
The above implies \eqref{eqn:figstar2},
\begin{align}
    \bar{s}_{\ell}\p{\pi, \s{z}_1^{\ell}} 
        &=   s_{\ell}\p{\pi, \s{z}_1^{\ell}} 
            - c \ell \pi\\
        &= \min \ppp{
                    g(\pi) \:,\: 
                    \E{
                        s_{\ell+1}(\Pi_{\ell + 1} , \calF_{\ell+1})
                        - c (\ell + 1) \pi
                        \vert
                        \Pi_{\ell} = \pi,
                        \calF_{\ell} = \s{z}_1^{\ell}
                    }
                    + c \pi
                }\\
        &=  \min \ppp{
                    g(\pi) \:,\: 
                    \E{
                        \bar{s}_{\ell+1}(\Pi_{\ell + 1} , \calF_{\ell+1})
                        \vert
                        \Pi_{\ell} = \pi,
                        \calF_{\ell} = \s{z}_1^{\ell}
                    }
                    + c \pi
                }.\label{eqn:figstar2}
\end{align}
Thus, at optimal stopping time $\s{T}^\star$, we have
\begin{align}
    \bar{s}_{\ell}\p{\pi, \s{z}_1^{\s{T}^\star}} 
        = g(\pi).
\end{align}
Furthermore, if $\bar{s}_{\ell}\p{\pi, \s{z}_1^{\ell}} = g(\pi)$ and $c_{\s{II}} < c_{\s{I}} (1 - \pi)$, then $\bar{s}_{\ell}\p{\pi, \s{z}_1^{\ell}} = c_{\s{II}} \pi$, and the information is declared as genuine; otherwise, it is declared as fake. Hence,
\begin{align}
    \pi_{\s{low}} (\s{z}_1^{\ell}) 
        &= \sup 
            \ppp{
                \pi \le \frac{c_{\s{I}}}{c_{\s{I}} + c_{\s{II}}} 
                : \bar{s}_{\ell} \p{\pi, \s{z}_1^{\ell}} 
                    = c_{\s{II}} \pi
            },
\end{align}
and
\begin{align}
    \pi_{\s{up}} (\s{z}_1^{\ell})  
        &= \inf
            \ppp{
                \frac{c_{\s{I}}}{c_{\s{I}} + c_{\s{II}}} \le \pi 
                : \bar{s}_{\ell}\p{\pi, \s{z}_1^{\ell}} 
                    = c_{\s{I}} (1 - \pi)
            }.
\end{align}

\subsection{Proof of Theorem~\ref{theorem:SPRT}}
If $\pi_{\s{low}} \le \pi \le \pi_{\s{up}}$, for all $\ell = 1, 2, \ldots$, then $\s{B}_{\s{low}}$ and $\s{B}_{\s{up}}$, defined in \eqref{eq:Y_low} and \eqref{eq:Y_up}, respectively, satisfy $0 < \s{B}_{\s{low}} \le 1 \le \s{B}_{\s{up}} < \infty$. 
From Theorem~\ref{theorem:OptimalStoppingTime}
the stopping time \eqref{eq:optimalStoppingTime} may be written 
\begin{align}
    \s{T}^\star = \inf 
            \{
                \ell\in\mathbb{N}: 
                \Pi_{\ell} \notin (\pi_{\s{low}}(\s{z}_1^{\ell}) , 
                \pi_{\s{up}}(\s{z}_1^{\ell}))\},
\end{align}
but because,
\begin{align}
    \Pi_{\ell}  
        &=  \P(\calH_1 \vert \calF_{\ell}) \\
        &=  \frac
                {\pi \P(\calF_{\ell} \vert \calH_1)}
                {
                    \pi \P(\calF_{\ell} \vert \calH_1)
                    + (1 - \pi) \P(\calF_{\ell} \vert \calH_0)
                } \\
        &=  \frac
                {\pi \Lambda_{\ell}}
                {
                    \pi \Lambda_{\ell}
                    + (1 - \pi) 
                },
\end{align}
\sloppy
we see that $\Pi_{\ell} \notin ( \pi_{\s{low}}(\s{z}_1^{\ell}), \pi_{\s{up}}(\s{z}_1^{\ell}))$ is equivalent to $\Lambda_{\ell} \notin (\s{B}_{\s{low}}(\s{z}_1^{\ell}), \s{B}_{\s{up}}(\s{z}_1^{\ell}))$ with $\s{B}_{\s{low}}(\s{z}_1^{\ell})$ and $\s{B}_{\s{up}}(\s{z}_1^{\ell})$, defined in \eqref{eq:Y_low} and \eqref{eq:Y_up}, respectively. Thus, the stopping time \eqref{eq:SPRT_stop_time} is equivalent to \eqref{eq:optimalStoppingTime}. Similarly, the decision rule \eqref{eq:decision_rule} may be written as,
\begin{align}
    \delta_{\s{T}^\star} = 
        \begin{cases}
            0,  &   \Pi_{\s{T}^\star} \le \pi_{\s{low}}(\s{z}_1^{\s{T}^\star}) \\
            1,  &   \Pi_{\s{T}^\star} > \pi_{\s{up}}(\s{z}_1^{\s{T}^\star}),
        \end{cases}
\end{align}
but again $\Pi_{\ell} \le \pi_{\s{low}}(\s{z}_1^{\ell})$ and $\Pi_{\ell} \ge \pi_{\s{up}}(\s{z}_1^{\ell})$ are equivalent to $\Lambda_{\ell} \le \s{B}_{\s{low}}(\s{z}_1^{\ell})$ and $\Lambda_{\ell}> \s{B}_{\s{up}}(\s{z}_1^{\ell})$, respectively, for all $\ell \ge 1$. Hence, the decision rule \eqref{eq:SPRT_decision_rule} is equivalent to \eqref{eq:decision_rule}.

\subsection{Proof of Proposition~\ref{prop:ErrorProb&Boundries}}
Recall that for each $\ell$, we fix $0 < \s{B}_{\s{low}} \le 1 \le \s{B}_{\s{up}} < \infty$. We let $\sigma_{\s{T}^\star} = \sigma(\s{z}_1, \ldots, \s{z}_{\s{T}^\star})$ be the $\sigma$-algebra. Consider the following chain of equations,  \begin{align}
    &\s{P}_{e,1}   =  \P_{\calH_0}{\delta_{\s{T}^\star} = 1}\\
        &=  \frac{1}{\s{Z}^{\s{T}^\star}}
            \sum\limits_{\calF_{\s{T}^\star} \in \sigma_{\s{T}^\star}}
                \P_{\calH_0}{\delta_{\s{T}^\star} = 1 \vert \calF_{\s{T}^\star}}\\
        &=  \frac{1}{\s{Z}^{\s{T}^\star}}
            \sum\limits_{\calF_{\s{T}^\star} \in \sigma_{\s{T}^\star}}   
                \P_{\calH_0}{\Lambda_{\s{T}^\star} \ge \s{B}_{\s{up}} \vert \calF_{\s{T}^\star}} 
                \label{eq:Tfinite}\\
        &=  \frac{1}{\s{Z}^{\s{T}^\star}}
            \sum\limits_{\calF_{\s{T}^\star} \in \sigma_{\s{T}^\star}} 
                \E{ \mathds{1}_{\{ \Lambda_{\s{T}^\star} \ge \s{B}_{\s{up}}\}}
                    \mathds{1}_{{\calH_0}}
                    \vert 
                    \calF_{\s{T}^\star}
                    }\\
        &=  \frac{1}{\s{Z}^{\s{T}^\star}}
            \sum\limits_{\calF_{\s{T}^\star} \in \sigma_{\s{T}^\star}} 
            \sum\limits_{\ell = 1}^{\infty}
                \E{ \mathds{1}_{\{ \Lambda_{\s{T}^\star} \ge \s{B}_{\s{up}}\}}
                    \mathds{1}_{{\calH_0}}
                    \vert 
                    \calF_{\s{T}^\star},
                    \s{T}^\star = \ell
                    }\nonumber\\
                    &\hspace{3cm}\cdot \P(\s{T}^\star = \ell)\\
        &\le \frac{1}{\s{Z}^{\s{T}^\star}}
            \sum\limits_{\calF_{\s{T}^\star} \in \sigma_{\s{T}^\star}} 
            \sum\limits_{\ell = 1}^{\infty}
                \frac{1}{\s{B}_{\s{up}}}
                \E{ \Lambda_{\s{T}^\star} 
                    \mathds{1}_{\{ \Lambda_{\s{T}^\star} \ge \s{B}_{\s{up}}\}}
                    \mathds{1}_{{\calH_0}}
                    \vert 
                    \calF_{\s{T}^\star},
                    \s{T}^\star = \ell
                    }\nonumber\\
                    &\hspace{3cm}\cdot \P(\s{T}^\star = \ell) \label{eq:BlessLambda}
                \\
    &= \frac{1}{\s{Z}^{\s{T}^\star}}
            \sum\limits_{\calF_{\s{T}^\star} \in \sigma_{\s{T}^\star}} 
            \sum\limits_{\ell = 1}^{\infty}
                \frac{1}{\s{B}_{\s{up}}}
                \E{ \mathds{1}_{\{ \Lambda_{\s{T}^\star} \ge \s{B}_{\s{up}}\}}
                    \mathds{1}_{{\calH_1}}
                    \vert 
                    \calF_{\s{T}^\star},
                    \s{T}^\star = \ell
                    }\nonumber\\
                    &\hspace{3cm}\cdot \P(\s{T}^\star = \ell)
                \label{eq:LambdaDef}\\
    &=  \frac{1}{\s{Z}^{\s{T}^\star}}
            \sum\limits_{\calF_{\s{T}^\star} \in \sigma_{\s{T}^\star}} 
                \frac{1}{\s{B}_{\s{up}}}
                \E{ \mathds{1}_{\{ \Lambda_{\s{T}^\star} \ge \s{B}_{\s{up}}\}}
                    \mathds{1}_{{\calH_1}}
                    \vert 
                    \calF_{\s{T}^\star}
                    }\\
    &=  \frac{\P_{\calH_1}(\Lambda_{\s{T}^\star} \ge \s{B}_{\s{up}})}{\s{B}_{\s{up}}}\\
    &=  \frac{1-\s{P}_{e,2}}{\s{B}_{\s{up}}},
\end{align}
where \eqref{eq:Tfinite} holds due to the optimally of the stopping time which assures that $\s{T}^\star$ is almost surely finite, \eqref{eq:BlessLambda} holds since $\Lambda_{\s{T}^\star} \ge \s{B}_{\s{up}}$, and finally \eqref{eq:LambdaDef} holds due to the fact the all events $\{\s{T}^\star = \ell\}$, $\{\calF_{\ell} \in \sigma_{\s{T}^\star}\}$ and $\{\Lambda_{\ell} \ge \s{B}_{\s{up}}\}$ are in $\calF_{\ell}$, and so,
\begin{align}
    &\E{ \Lambda_{\ell} 
        \mathds{1}_{\{\Lambda_{\ell} \ge \s{B}_{\s{up}}\}}
        \mathds{1}_{{\calH_0}}
        \vert 
        \calF_{\s{T}^\star}, \s{T}^\star = \ell}\nonumber\\
    &=  \int\limits_{
                \ppp{   \Lambda_{\ell} \ge \s{B}_{\s{up}},
                        \calF_{\s{T}^\star} \in \sigma_{\s{T}^\star},  
                        \s{T}^\star = \ell}    } 
            \Lambda_{\ell} d\P_0 \\
    &=  \int\limits_{
                \ppp{   \Lambda_{\ell} \ge \s{B}_{\s{up}}, 
                        \calF_{\s{T}^\star} \in \sigma_{\s{T}^\star} \s{T}^\star = \ell} } 
                        d\P_1 \\
    &= \E{  \mathds{1}_{\{\Lambda_{\ell} \ge \s{B}_{\s{up}}\}}
            \mathds{1}_{{\calH_1}}
            \vert \calF_{\s{T}^\star}, \s{T}^\star = \ell}.
\end{align}
Finally, a similar argument gives $\s{P}_{e,2} \le \s{B}_{\s{low}} (1-\s{P}_{e,1})$, which concludes the proof.

\section{Conclusion} \label{section:Conclusions}
This paper introduces a quickest misinformation detection algorithm based on a realistic probabilistic model of information propagation through a social media platform. The problem is formulated and solved as an optimal stopping problem that minimizes the combination of the error probability and the stopping time. Our numerical results with a real-world data demonstrate that our algorithm outperforms state-of-the-art early misinformation detection algorithms. As an interesting direction for future research, while in this work, we considered a hard decision problem between two possible hypotheses (genuine or fake), it is more reasonable and robust to consider a softer decision problem, with multiple hypotheses, or even a sequential estimation problem, where the parameter to be estimated reflects the level of genuineness/fakeness. 




\bibliographystyle{abbrv}
\bibliography{bibfile}




\end{document}